\documentclass[journal]{IEEEtran}
\usepackage{cite}
\usepackage{graphicx}
\usepackage{url}
\usepackage{amsmath}
\usepackage[caption = false, font=footnotesize]{subfig}
\usepackage{float}
\usepackage{booktabs} % Allows the use of \toprule, \midrule and \bottomrule in tables for horizontal lines
\usepackage{multirow}
\usepackage{array}
\usepackage{algorithm}
\usepackage{algpseudocode}
\usepackage{bbm}
\usepackage{dsfont}
\usepackage{tabu}
\newcolumntype{L}[1]{>{\raggedright\let\newline\\\arraybackslash\hspace{0pt}}m{#1}}
\newcolumntype{C}[1]{>{\centering\let\newline\\\arraybackslash\hspace{0pt}}m{#1}}
\newcolumntype{R}[1]{>{\raggedleft\let\newline\\\arraybackslash\hspace{0pt}}m{#1}}

\ifCLASSINFOpdf
\else
\fi
\begin{document}
\title{dipIQ: Blind Image Quality Assessment by Learning-to-Rank Discriminable Image Pairs}

\author{Kede~Ma,~\IEEEmembership{Student~Member,~IEEE,}
        Wentao~Liu,~\IEEEmembership{Student~Member,~IEEE,}
        Tongliang~Liu,\\
        Zhou~Wang,~\IEEEmembership{Fellow,~IEEE,}
        and~Dacheng~Tao,~\IEEEmembership{Fellow,~IEEE}% <-this % stops a space
\thanks{K. Ma, W. Liu, and Z. Wang are with the Department of Electrical and Computer Engineering, University of Waterloo, Waterloo, ON N2L 3G1, Canada (e-mail: \{k29ma, w238liu, zhou.wang\}@uwaterloo.ca).}%
\thanks{T. Liu and D. Tao are with the UBTech Sydney Artificial Intelligence Institute and the School of Information Technologies in the Faculty of Engineering and Information Technologies at The University of Sydney, J12 Cleveland St, Darlington, NSW 2008, Australia (email: \{tongliang.liu, dacheng.tao\}@sydney.edu.au).}
\thanks{\textcopyright~2017 IEEE. Personal use of this material is permitted. Permission from IEEE must be obtained for all other uses, in any current or future media, including reprinting/republishing this material for advertising or promotional purposes, creating new collective works, for resale or redistribution to servers or lists, or reuse of any copyrighted component of this work in other works.}
% <-this % stops a space
}


\markboth{}%
{Shell \MakeLowercase{\textit{et al.}}: Bare Demo of IEEEtran.cls for Journals}



\maketitle

\begin{abstract}
 Objective assessment of image quality is fundamentally important in many image processing tasks. In this work, we focus on learning blind image quality assessment (BIQA) models which predict the quality of a digital image with no access to its original pristine-quality counterpart as reference. One of the biggest challenges in learning BIQA models is the conflict between the gigantic image space (which is in the dimension of the number of image pixels) and the extremely limited reliable ground truth data for training. Such data are typically collected via subjective testing, which is cumbersome, slow, and expensive. Here we first show that a vast amount of reliable training data in the form of quality-discriminable image pairs (DIP) can be obtained automatically at low cost by exploiting large-scale databases with diverse image content. We then learn an opinion-unaware BIQA (OU-BIQA, meaning that no subjective opinions are used for training) model using RankNet, a pairwise learning-to-rank (L2R) algorithm, from millions of DIPs, each associated with a perceptual uncertainty level, leading to a DIP inferred quality (dipIQ) index. Extensive experiments on four benchmark IQA databases demonstrate that dipIQ outperforms state-of-the-art OU-BIQA models. The robustness of dipIQ is also significantly improved as confirmed  by the group MAximum Differentiation (gMAD) competition method. Furthermore, we extend the proposed framework by learning models with ListNet (a listwise L2R algorithm) on quality-discriminable image lists (DIL). The resulting DIL Inferred Quality (dilIQ) index achieves an additional performance gain.

 %The proposed pairwise and listwise learning schemes for IQA are generic, allowing new and flexible image pairs and lists generation engines, and more advanced pairwise and listwise L2R algorithms to be directly incorporated to improve the performance.

\end{abstract}


% \fi
\IEEEpeerreviewmaketitle

\section{Introduction}
\IEEEPARstart{O}{bjectively} assessing image quality is of fundamental importance  due in part to the massive expansion of online image volume. Objective image quality assessment (IQA) has become an active research topic over the last decade, with a large variety of IQA models proposed~\cite{wu2005digital,wang2006modern}. They can be categorized into full-reference models (FR, where the reference image is fully available when evaluating a distorted image)~\cite{daly1992visible}, reduced-reference models (RR, where only partial information about the reference image is available)~\cite{wang2006quality}, and blind/no-reference models (NR, where the reference image is not accessible)~\cite{wang2011reduced}. In many real-world applications, reference images are unavailable, making blind IQA (BIQA) models highly desirable in practice.

Many BIQA models are developed by supervised learning~\cite{moorthy2010two,saad2012blind,mittal2012no,ye2012unsupervised,moorthy2011blind,wu2015a,xue2014blind,gu2015using,wu2014blind} and share a common two-stage structure: 1) perception- and/or distortion-relevant features (denoted by $\bf x$) are extracted from the test image; and 2) a quality prediction function $f({\bf x})$ is learned by statistical machine learning algorithms. The performance and robustness of these approaches rely heavily on the quality and quantity of the ground truth data for training. The most common type of ground truth data is in the form of the mean opinion score (MOS), which is the average of quality ratings given by multiple subjects. Therefore, these models are often referred to as opinion-aware BIQA (OA-BIQA) models and may
 %This learning scheme is the same as in pointwise learning-to-rank (L2R) approaches, where the degree of importance of each instance is usually used for training and is also what we intend to predict\footnote{In the context of document retrieval, ``the degree of importance of each instance'' denotes the degree of relevance of each document associated with a certain query, while in the context of IQA, it means the degree of perceived quality of each test image.}.Mean opinion scores (MOSs) for training aim to make them opinion aware (OA), but
 incur the following drawbacks. First, collecting MOS via subjective testing is slow, cumbersome, and expensive. As a result, even the largest publicly available IQA database, TID2013~\cite{Ponomarenko201557}, provides only $3,000$ images with MOSs. This limited number of training images is deemed extremely sparsely distributed in the entire image space, whose dimension equals the number of pixels and is typically in the order of millions. As such, the generalizability of BIQA models learned from small training samples is questionable on real-world images. Second, among thousands of sample images, only a few dozen source reference images can be included, considering the combinations of reference images, distortion types and  levels. For example, the TID2013 database~\cite{Ponomarenko201557} includes $25$ source images only. It is extremely unlikely that this limited number of reference images sufficiently represent the variations that exist in real-world images. Third, since these BIQA models are trained with individual images to make independent quality predictions, the cost function is blind to the relative perceptual order between images. As a result, the learned models are weak at ordering images with respect to their perceptual quality.

In this paper, we show that a vast amount of reliable training data in the form of so-called quality-discriminable image pairs (DIP) can be generated by exploiting large-scale databases with diverse image content. Each  DIP is associated with a perceptual uncertainty measure to indicate the confidence level of its quality discriminability. We show that such DIPs can be generated at very low cost without resorting to subjective testing. We then employ RankNet~\cite{Burges05learningto}, a neural network-based pairwise learning-to-rank (L2R) algorithm~\cite{liu2009learning,hang2011short}, to learn an opinion-unaware BIQA (OU-BIQA, meaning that no subjective opinions are used for training) model by incorporating the uncertainty measure into the loss function. Extensive experiments on four benchmark IQA databases demonstrate that the DIP inferred quality (dipIQ) indices significantly outperform previous OU-BIQA models. We also conduct another set of experiments in which we train the dipIQ indices using different feature representations as inputs and compare them with OA-BIQA models using the same representations. The generalizability and robustness of dipIQ are improved across all four IQA databases and verified by the group MAximum Differentiation (gMAD) competition method~\cite{ma2016group}, which examines image pairs optimally selected from the Waterloo Exploration Database~\cite{ma2016waterloo}. Furthermore, we extend the  proposed pairwise L2R approach for OU-BIQA to a listwise L2R one by evoking ListNet~\cite{cao2007learning} (a listwise L2R extension of RankNet~\cite{Burges05learningto}) and transforming DIPs to quality-discriminable image lists (DIL) for training. The resulting DIL inferred quality (dilIQ) index leads to an additional performance gain.

The remainder of the paper is organized as follows. BIQA models and typical L2R algorithms are reviewed and categorized in Section~\ref{sec:rw}. The proposed dipIQ approach is introduced in Section~\ref{sec:pl}. Experimental results using dipIQ on four benchmark IQA databases compared with state-of-the-art BIQA models are presented in Section~\ref{sec:exp}, followed by an extension to the dilIQ model in Section~\ref{sec:ListNet}. We conclude the paper in Section~\ref{sec:con}.

\section{Related Work}
\label{sec:rw}
We first review existing BIQA models according to their two-stage structure: feature extraction and quality prediction model learning. We then review typical L2R algorithms. Details of RankNet~\cite{Burges05learningto} are provided in Section~\ref{sec:pl}.

\subsection{Existing BIQA Models}
From the feature extraction point of view, three types of knowledge can be exploited to craft useful features for BIQA. The first is knowledge about our visual world that summarizes the statistical regularities of undistorted images. The second is knowledge about degradation, which can then be explicitly taken into account to build features for particular artifacts, such as blocking~\cite{wu1997generalized,Wang2000Blind,liu2002efficient}, blurring~\cite{tong2004blur,wang2003local,zhu2009no} and ringing~\cite{ouguz1998image,sheikh2005no,liu2010no}. The third is knowledge of the human visual system (HVS)~\cite{wandell1995foundations}, namely perceptual models derived from visual physiological and psychophysical studies~\cite{hubel1962receptive,heeger1992normalization,field1994goal,geisler2002bayesian}. Natural scene statistics (NSS), which seek to capture the natural statistical behavior of images, embody the three-fold modeling in a rather elegant way~\cite{wang2011reduced}. NSS can be extracted directly in the spatial domain or in transform domains such as DFT, DCT, and  wavelets~\cite{simoncelli1992shiftable,mallat1989theory}.

In the spatial domain, edges are presumably the most important image features. The edge spread can be used to detect blurring~\cite{li2002blind,marziliano2004perceptual}, and the intensity variance in smooth regions close to edges can indicate ringing artifacts~\cite{ouguz1998image}. Step edge detectors that operate at $8 \times8$ block boundaries measure the severity of discontinuities caused by JPEG compression~\cite{wu1997generalized}. The sample entropy of intensity histograms is used to identify image anisotropy~\cite{li2011blind,fang2015no}. The responses of image gradients and the Laplacian of Gaussian operators are jointly modeled to describe the destruction of statistical naturalness of images~\cite{xue2014blind}. The singular value decomposition of local image gradient matrices may provide a quantitative measure of image content~\cite{zhu2010automatic}. Mean-subtracted and contrast-normalized pixel value statistics have also been modeled using a generalized Gaussian distribution (GGD)~\cite{mittal2012no,mittal2013making,mittal2012blind,ye2013real}, inspired by the adaptive gain control mechanism seen in neurons~\cite{heeger1992normalization}.

Statistical modeling in the wavelet domain resembles the early visual system~\cite{hubel1962receptive}, and natural images exhibit statistical regularities in the wavelet space. Specifically, it is widely acknowledged that the marginal distribution of wavelet coefficients of a natural image (regardless of content) has a sharp peak near zero and heavier than Gaussian tails. Therefore, statistics of raw~\cite{wang2005reduced,wang2006quality,moorthy2010two,hou2015blind} and normalized~\cite{li2009reduced,rehman2012reduced} wavelet coefficients, and wavelet coefficient correlations in the neighborhood~\cite{sheikh2005no,moorthy2011blind,tang2011learning,tang2014blind,ye2012no} can be individually or jointly modeled as image naturalness measurements. The phase information of wavelet coefficients, for example expressed as the  local phase coherence, is exploited to describe the perception of blur~\cite{wang2003local} and sharpness~\cite{hassen2013image}.

In the DFT domain, blur kernels can be efficiently estimated~\cite{tang2011learning,xu2010two,tang2014blind} to quantify the degree of image blurring. The regular peaks at feature frequencies can be used to identity  blocking artifacts~\cite{Wang2000Blind,wang2002no}. Moreover, it is generally hypothesized that most perceptual information in an image is stored in the Fourier phase rather than the Fourier amplitude~\cite{huang1975importance,oppenheim1981importance}. Phase congruency~\cite{kovesi1999image} is such a feature that identifies perceptually significant image features at spatial locations where Fourier components are maximally in-phase~\cite{li2011blind}.

In the DCT domain, blocking artifacts can be identified in a shifted $8\times 8$ block~\cite{liu2002efficient}. The ratio of AC coefficients to DC components can be interpreted as a measure of local contrast~\cite{saad2010dct}. The kurtosis of AC coefficients can be used to quantify the structure statistics. In addition, AC coefficients can also be jointly modeled using a GGD~\cite{saad2012blind}.

There is a growing interest in learning features for BIQA. Ye {\em et al.} learned quality filters on image patches using K-means clustering and adopted filter responses as features~\cite{ye2012unsupervised}. They then took one step further by supervised filter learning~\cite{ye2013real}. Xue {\em et al.}~\cite{xue2013learning} proposed a quality-aware clustering scheme on the high frequencies of raw patches, guided by an FR-IQA measure~\cite{zhang2011fsim}. Kang {\em et al.} investigated a convolutional neural network to jointly learn features and nonlinear mappings for BIQA~\cite{kang2014convolutional}.

%As a final note, the dimension of the feature vector ${\bf x}$ extracted from the test image may be extremely high, which poses a challenge to  traditional machine learning algorithms. As a result, dimensionality reduction techniques, such as PCA~\cite{vidal2005generalized} may first be performed in order to reduce $\bf x$ to a lower dimension.

From the model learning perspective, SVR~\cite{cortes1995support,scholkopf2000new} is the most commonly used tool to learn  $f({\bf x})$ for BIQA~\cite{moorthy2010two,moorthy2011blind,ye2012no,ye2012unsupervised,ye2013real,xue2014blind}. The capabilities of neural networks to pre-train a model without labels and to easily scale up have also been exploited for this purpose~\cite{li2011blind,kang2014convolutional,tang2014blind,hou2015blind}. Another typical quality regression is the example-based method, which predicts the test image quality score using the weighted average of training image quality scores, where the weight encodes the perceptual similarity between the test and training images~\cite{ye2012no,xue2013learning,wu2014blind}. Saad {\em et al.} jointly modeled $\bf x$ and MOS using a multivariate Gaussian distribution and performed prediction by maximizing the conditional probability $P({\bf x}|\mathrm{MOS})$~\cite{saad2010dct,saad2012blind}. Similar probabilistic modeling strategies have been investigated~\cite{mittal2013making,zhang2015feature}. Pairwise L2R algorithms have also been used to learn BIQA models~\cite{xu2014rank,gao2015learning}. However, in these methods, DIP generation relies solely on MOS availability, which limits the number of DIPs produced. Moreover, their performance is inferior to that of existing BIQA methods. Other advanced learning algorithms include topic modeling~\cite{hofmann2001unsupervised}, Gaussian process~\cite{tang2014blind}, and multi-kernel learning~\cite{gao2013universal,gao2015learning}.

\subsection{Existing L2R Algorithms}
Existing L2R algorithms can be broadly classified into three categories based on the training data format and loss function: pointwise, pairwise, and listwise approaches. An excellent survey of L2R algorithms can be found in~\cite{liu2009learning}. Here we only provide a brief overview.

Pointwise approaches assume that each instance's importance degree is known. The loss function usually examines the prediction accuracy of each individual instance. In an early attempt on L2R, Fuhr~\cite{Fuhr1989Optimum} adopted a linear regression with a polynomial feature expansion to learn the score function $f(\bf x)$. Cossock and Zhang~\cite{Cossock2006Subset} utilized a similar formulation with some theoretical justifications for the use of the least squares loss function. Nallapati~\cite{Nallapati2004Discriminative} formulated L2R as a classification problem and investigated the use of maximum entropy and support vector machines (SVMs) to classify each instance into two classes---relevant or irrelevant. Ordinal regression-based pointwise L2R algorithms have also been proposed such as PRanking~\cite{Crammer2002Pranking} and SVM-based large margin principles~\cite{shashua2002ranking}.

%Pointwise approaches can only deliver sub-optimal solutions because relative order between instances cannot be naturally considered in the learning process. Furthermore, the evaluation measures for ranking cannot be well considered and directly optimized by pointwise approaches~\cite{liu2009learning}.

Pairwise approaches assume that the relative order between two instances is known or can be inferred from other ground truth formats. The goal is to minimize the number of misclassified instance pairs. In the extreme case, if all instance pairs are correctly classified, they will be correctly
ranked~\cite{liu2009learning}. In RankSVM~\cite{Joachims2002Optimizing}, Joachims creatively generated training pairs from clickthrough data and reformulated SVM to learn the score function $f({\bf x})$ from instance pairs. Proposed in 2005, RankNet~\cite{Burges05learningto} was probably the first L2R algorithm used by commercial search engines, which had a typical neural network with a weight-sharing scheme forming its skeleton. %The cross entropy loss function defined on the pair with the help of $f$ has a desirable probability interpretation. 
Tsai {\em et al.}~\cite{Tsai2007FRank} replaced RankNet's loss function~\cite{Burges05learningto} with a fidelity loss originating from quantum physics. In this paper, RankNet is adopted as the default pairwise L2R algorithm to learn OU-BIQA models for reasons that will be described later. RankBoost~\cite{Freund2003An} is another well-known pairwise L2R algorithm based on AdaBoost~\cite{Freund1995A} with an exponential loss.

Listwise approaches provide the opportunity to directly optimize ranking performance criteria~\cite{liu2009learning}. Representative algorithms include SoftRank~\cite{taylor2008softrank}, $\textrm{SVM}^{map}$~\cite{yue2007support}, and RankGP~\cite{yeh2007learning}. Another subset of listwise approaches choose to optimize listwise ranking losses. For example, as a direct extension of RankNet, ListNet~\cite{cao2007learning} duplicates RankNet's structure to accommodate an instance list as input and optimizes a ranking loss based on the permutation probability distribution~\cite{cao2007learning}. In this paper, we also employ ListNet to learn OU-BIQA models as an extension of the proposed pairwise L2R approach.

 % With the pairwise approach, the relative perceptual order between two images can be better modeled. With the listwise approach, the rank information among images in a given list is visible to the learning process.

\begin{figure*}[t]
  \centering
  \includegraphics[width=1 \linewidth]{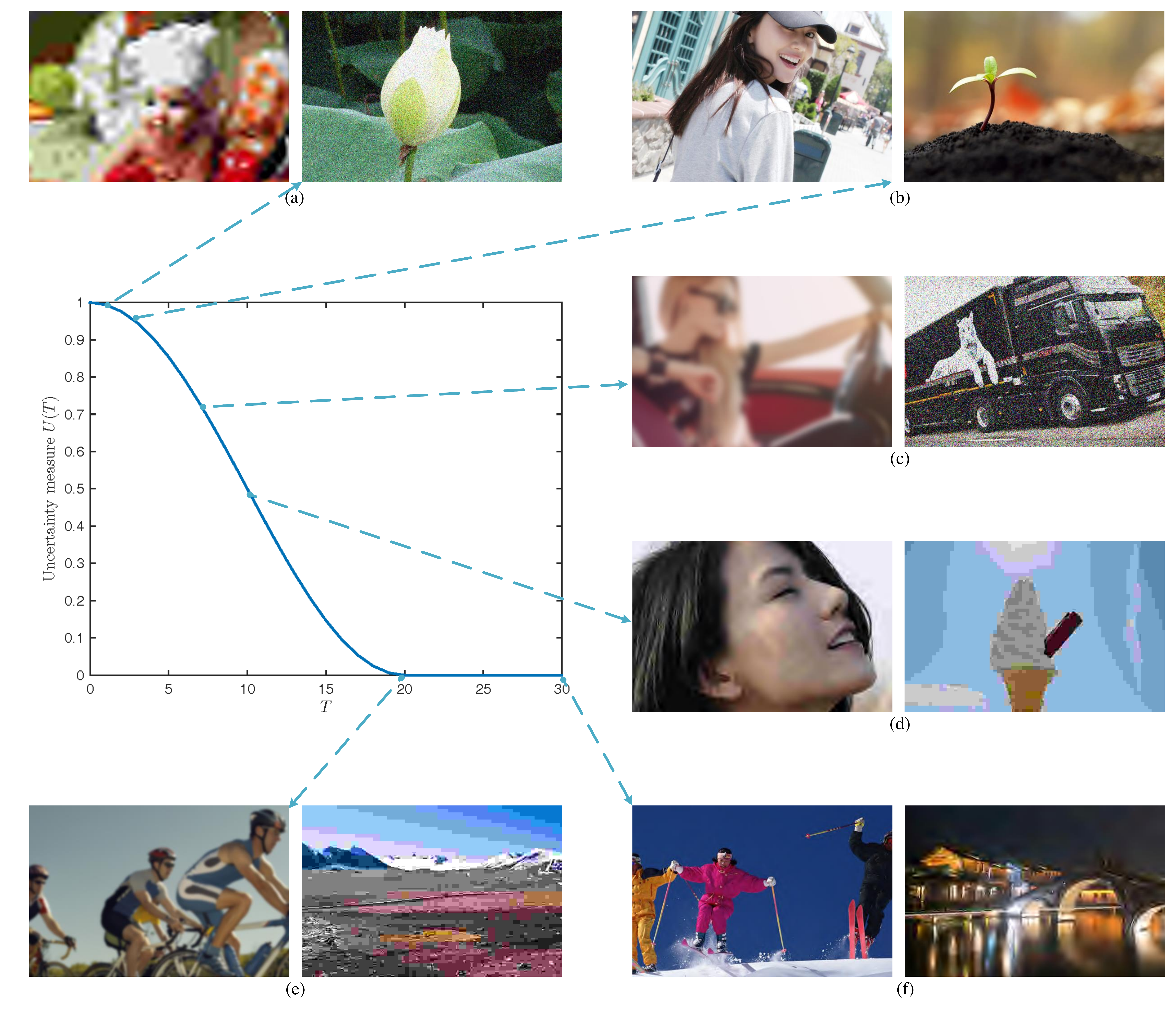}\\
  \caption{Illustration of the perceptual uncertainty of quality discriminability of DIPs as a function of $T$. The left images of all DIPs have better quality in terms of the three FR-IQA models with $T>0$. However, the quality discriminability differs significantly. All images are originated from the $700$ training images and cropped for better visibility.}\label{fig:uncertainty}
\end{figure*}

\section{Proposed Pairwise L2R Approach for OU-BIQA}\label{sec:pl}
In this section, we elaborate the proposed pairwise L2R approach to learn OU-BIQA models. First, we propose an automatic DIP generation engine. Each DIP is associated with an uncertainty measure to quantify the confidence level of its quality discriminability.
Second, we detail RankNet~\cite{Burges05learningto} and extend its capability to learn from the generated DIPs with uncertainty.

\subsection{DIP Generation}\label{subsec:DIP}
Our automatic DIP generation engine is described as follows. We first choose three best-trusted FR-IQA models, namely MS-SSIM~\cite{wang2003multiscale}, VIF~\cite{sheikh2006image}, and GSMD~\cite{xue2014gradient}. A logistic nonlinear function suggested in~\cite{sheikh2006statistical} is adopted to map predictions of the three models to the MOS scale of the LIVE database~\cite{LIVE}. After that, the score range of the three models roughly spans $[0, 100]$, where higher values indicate better perceptual quality. We associate each candidate image pair with a nonnegative $T$, which is equal to the smallest score difference of the three FR models. 
Intuitively, the perceptual uncertainty level of quality discriminability should decrease monotonically with the increase of $T$. By varying $T$, we can generate DIPs with different uncertainty levels. To quantify the level of uncertainty, we employ a raised-cosine function given by
%\begin{equation}
%\label{eq:gcdf} U(T) =
%\frac{1}{\sqrt{2\pi}\sigma}\int_{-\infty}^{T_c - T}\mathrm{exp}\left[-\frac{(\omega - \mu)^2}{2\sigma^2}\right]d\omega\,,
%\end{equation}
\begin{equation}
\label{eq:gcdf}
     U(T) =
   \begin{cases}
   \frac{1}{2}\left(1 + \cos\left(\frac{\pi T}{T_c}\right)\right) &\mbox{if $T \le T_c$}\\
  0 &\mbox{otherwise }\,,
   \end{cases}
\end{equation}
where $U(T)$ lies in $[0, 1]$, with a higher value indicating a greater degree of uncertainty and $T_c$ is a constant, above which the uncertainty goes to zero. In the current implementation, we set $T_c = 20$, whose legitimacy can be validated from two sources. First, the average standard deviation of MOSs on LIVE is around $9$, which is approximately half of $T_c$, therefore guaranteeing the perceived discriminability of two images. Second, based on the subjective experiments conducted by Gao {\em et al.}~\cite{gao2015learning} on LIVE, the consistency between subjects on the relative quality of one pair increases with the absolute difference and, when it is larger than $20$, the consistency approaches $100\%$. Fig.~\ref{fig:uncertainty} shows the shape of the uncertainty function as a function of $T$ and some representative DIPs, where the left images have better quality in terms of the three chosen FR-IQA models with $T>0$. All the shown DIPs are generated from the training image set that will be described later. It is clear that setting $T$  close to zero produces the highest level of uncertainty of quality discriminability. Careful inspection of Fig.~\ref{fig:uncertainty}(a) and Fig.~\ref{fig:uncertainty}(b) reveals that the uncertainty manifests itself in two ways. First, the right image in Fig.~\ref{fig:uncertainty}(a) has better perceived quality to many human observers compared with the left one, which disagrees with the three FR-IQA models. Second, both images in Fig.~\ref{fig:uncertainty}(b) have  distortions that are barely perceived by the human eye. In other words, they have very similar perceptual quality. The perceptual uncertainty generally decreases if $T$ increases and when $T>20$, the DIP is clearly discriminable, further justifying the selection of $T_c = 20$.

\subsection{RankNet~\cite{Burges05learningto}}\label{subsec:ranknet}
\begin{figure}[t]
  \centering
  \includegraphics[width=1 \linewidth]{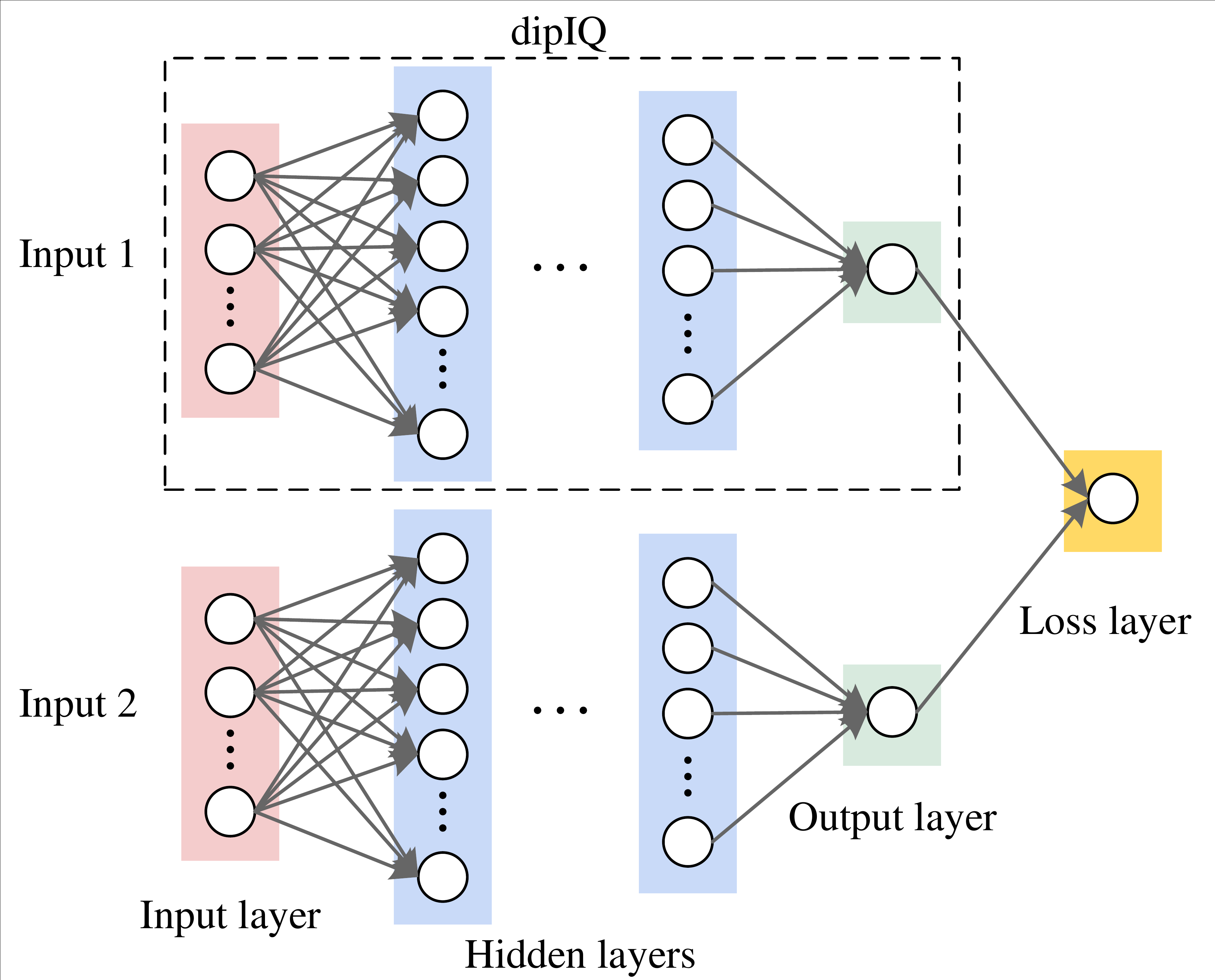}\\
  \caption{The architecture of dipIQ based on RankNet~\cite{Burges05learningto}.}\label{fig:ranknet}
\end{figure}

Given a number of DIPs, a pairwise L2R algorithm would make use of their perceptual order to learn quality models while taking the inherent perceptual uncertainty into account. Here, we revisit RankNet~\cite{Burges05learningto}, a pairwise L2R algorithm that was the first of its kind used by commercial search engines~\cite{liu2009learning}. We extend it to learn from DIPs associated with uncertainty.  Fig.~\ref{fig:ranknet} shows RankNet's architecture, which is based on classical neural networks and has two parallel streams  to accommodate a pair of inputs. The two-stream weights are shared, which is achieved by using the same initializations and the same gradients during backpropagation~\cite{Burges05learningto}. The quality prediction function $f({\bf x})$, namely the dipIQ index, is implemented by one of the streams, and the loss function is defined on a pair of images with the help of $f$.
Specifically, let $f({\bf x}_i)$ and $f({\bf x}_j)$ be the output of the first and second streams,  whose difference is converted to a probability using
\begin{equation}\label{eq:p}
P_{ij}(f) = \frac{\exp\left(f({\bf x}_i) - f({\bf x}_j)\right)}{1 + \exp\left(f({\bf x}_i) - f({\bf x}_j)\right)}\,,
\end{equation}
based on which we define the cross entropy loss as
\begin{equation}\label{eq:ce}
\begin{split}
&L(f;{\bf x}_i,{\bf x}_j, \bar{P}_{ij}) = -\bar{P}_{ij}\log P_{ij} - (1 - \bar{P}_{ij})\log(1 - P_{ij})\\
&= -\bar{P}_{ij}\left(f({\bf x}_i) - f({\bf x}_j)\right) + \log\left(1+\exp\left(f({\bf x}_i) - f({\bf x}_j)\right)\right)\,,
\end{split}
\end{equation}
where $\bar{P}_{ij}$ is the ground truth label associated with the training pair, consisting of the $i$-th and $j$-th images. In the case of DIPs described in the Section~\ref{subsec:DIP}, $\bar{P}_{ij}$ is always $0$ or $1$, indicating that the quality of the $i$-th image is worse or better than the $j$-th one. Within the mini-batch stochastic gradient minimization framework, we define the batch-level loss function using the perceptual uncertainty of each DIP as a weighting factor
\begin{equation}\label{eq:miniloss}
L_b(f) = \sum_{\langle i, j \rangle \in \mathcal{B}}(1 - U_{ij})L(f;{\bf x}_i,{\bf x}_j, \bar{P}_{ij})\,,
\end{equation}
where $\mathcal{B}$ is the batch containing the DIP indices currently being trained. As Eq.~(\ref{eq:miniloss}) makes clear, DIPs with higher uncertainty contribute less to the overall loss.
With some derivations, we obtain the gradient of $L_b$ with respect to the model parameters collectively denoted by $\bf w$ as follows
\begin{equation}\label{eq:gradient}
\begin{split}
\frac{\partial L_b(f) }{\partial \bf w}=& \sum_{\langle i, j \rangle \in \mathcal{B}}\left(-\bar{P}_{ij}+\frac{\exp\left(f({\bf x}_i) - f({\bf x}_j)\right)}{1+\exp\left(f({\bf x}_i) - f({\bf x}_j)\right)}\right)\\
& \bigg(1 - U_{ij} \bigg)\left(\frac{\partial f({\bf x}_i)}{\partial \bf w} - \frac{\partial f({\bf x}_j)}{\partial \bf w}\right)
\,.
\end{split}
\end{equation}
In the case of a linear dipIQ containing no hidden layers and no nonlinear activations, Eq.~(\ref{eq:ce}) is reduced to
\begin{equation}\label{eq:logr}
\begin{split}
L({\bf w}; {\bf x}_i, {\bf x}_j,\bar{P}_{ij}) = &-\bar{P}_{ij}\left({\bf w}^T({\bf x}_i - {\bf x}_j)\right)\\
&+\log(1+\exp({\bf w}^T({\bf x}_i - {\bf x}_j))\,,
\end{split}
\end{equation}
which is easily recognized as logistic regression. The convexity of Eq.~(\ref{eq:logr}) ensures the global optimality of the solution.  We investigate both linear and nonlinear dipIQ cases with the cross entropy as loss. In fact, any probability distribution measures can be adopted as alternatives. For example, Tsai {\em et al.}~\cite{Tsai2007FRank} proposed a fidelity loss measure from quantum physics. We find in our experiments that the fidelity loss impairs performance, so we use the cross entropy loss throughout the paper.

We select RankNet~\cite{Burges05learningto} as our first choice of pairwise L2R algorithm for two reasons. First, it is capable of handling a large number (millions) of training samples using stochastic or mini-batch gradient descent algorithms. By contrast, the training of other pairwise L2R methods such as RankSVM~\cite{Joachims2002Optimizing}, even with a linear kernel, is painfully slow. Second, since RankNet~\cite{Burges05learningto} embodies classical neural network architectures, we embrace the latest advances in training deep neural networks~\cite{hinton2006fast,krizhevsky2012imagenet} and can easily upscale the network by adding more hidden layers to learn powerful nonlinear quality prediction functions.

\section{Experiments}\label{sec:exp}
In this section, we first provide thorough implementation details of RankNet~\cite{Burges05learningto} to learn OU-BIQA models. We then describe the experimental protocol based on which a fair comparison is conducted between dipIQ and state-of-the-art BIQA models.  After that, we discuss how to extend the proposed pairwise L2R approach for OU-BIQA to a listwise one that could possibly boost the performance.

\begin{figure*}[t]
  \centering
  \includegraphics[width=1 \linewidth]{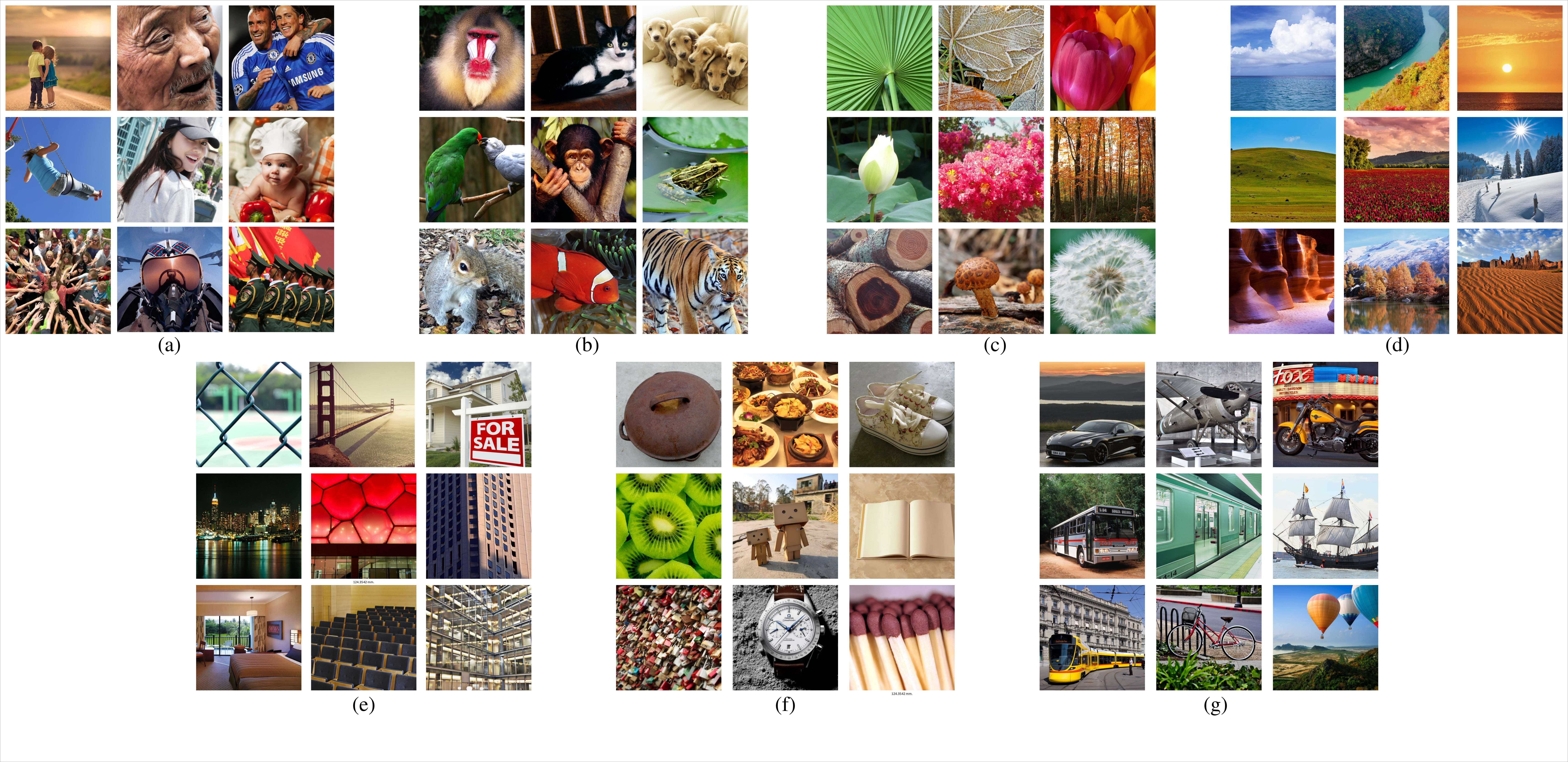}\\
  \caption{Sample source images in the training set. (a) Human. (b) Animal. (c) Plant. (d) Landscape. (e) Cityscape. (f) Still-life. (g) Transportation. All images are cropped for better visibility.}\label{fig:trainImg}
\end{figure*}

\subsection{Implementation Details}\label{subsec:id}
\subsubsection{Training Set Construction}\label{subsubsec:tsc}
We collect $840$ high quality and high resolution natural images to represent scenes we see in the real-world. They can be roughly clustered into seven groups: human, animal, plant, landscape, cityscape, still-life, and transportation. Sample source images are shown in Fig.~\ref{fig:trainImg}. We preprocess each source image by down-sampling it using a bicubic kernel so that the maximum height or width is $768$. Following the procedures described in~\cite{ma2016group}, we add four distortion types, namely JPEG and JPEG2000 (JP2K) compression, white Gaussian noise contamination (WN), and Gaussian blur (BLUR), each with five distortion levels. As a result, our training set consists of $17,640$ test images, with $840$ source and $16,800$ distorted images. We randomly hold out $140$ source images and their corresponding distorted images and use them as the validation set. For the rest $14,700$ images, we adopt the proposed DIP generation engine to produce more than $80$ million DIPs, which constitute our training set.
\subsubsection{Base Feature}\label{subsubsec:bf}
We adopt CORNIA features~\cite{ye2012unsupervised} to represent test images because they appear to be  highly competitive in a recent gMAD competition on the Waterloo Exploration Database~\cite{ma2016group}. In addition, a top performing OU-BIQA model, BLISS~\cite{ye2014beyond}, also chooses CORNIA features as input and trains on synthetic scores. As such, we offer a fair testing bed to compare dipIQ learned by a pairwise L2R approach (RankNet~\cite{Burges05learningto}) against BLISS~\cite{ye2014beyond} learned by a regression method (SVR).

\subsubsection{RankNet Instantiation}\label{subsubsec:ri}
We investigate both linear and nonlinear dipIQ models, denoted by dipIQ$^*$ and dipIQ, respectively. The input dimension to RankNet is $20,000$, equaling the feature dimension in CORNIA~\cite{ye2012unsupervised}. The loss layer is implemented by the cross entropy function in Eq.~(\ref{eq:ce}). For dipIQ$^*$, the input layer is directly connected to the output layer without adding hidden layers or going through nonlinear transforms. The use of the cross entropy loss ensures the convexity of the optimization problem. For dipIQ, we add $3$ hidden layers, which have a $256$ - $128$ - $3$ structure. All layers are fully connected, followed by rectified linear units (ReLU)~\cite{nair2010rectified} as nonlinearity activations. We choose the node number of the third hidden layer to be $3$ so that we can visualize the three-dimensional embedding of test images. Other choices are
somewhat ad-hoc, and a more careful exploration of alternative architectures could potentially lead to significant performance improvements.

The RankNet training procedure generally follows Simonyan and Zisserman~\cite{Simonyan2015Very}. Specifically, the training is carried out by optimizing the cross entropy function using mini-batch gradient descent with momentum. The weights of the two streams in RankNet are shared. The batch size is set to $512$, and momentum to $0.9$. The training is regularized by weight decay (the $L_2$ penalty multiplier set to $5\times10^{-4}$).  The learning rate is fixed to $10^{-4}$.
Since we have a plenty of DIPs (more than $80$ million) for training, each DIP is exposed to the learning algorithm once and only once. The learning stops when the entire set of DIPs have been swept. The weights that achieve the lowest validation set loss are used for testing.

\subsection{Experimental Protocol}\label{subsec:ep}
\subsubsection{Databases}\label{subsubsec:databases}
Four IQA databases are used to compare dipIQ  with state-of-the-art BIQA measures. They are LIVE~\cite{LIVE}, CSIQ~\cite{larson:011006}, TID2013~\cite{Ponomarenko201557} and Waterloo Exploration Database~\cite{ma2016waterloo}. The first three are small subject-rated IQA databases that are widely adopted to benchmark objective IQA models. Each test image is associated with an MOS to represent its perceptual quality. In our experiments, we only consider distortion types that are shared by all four databases, namely JP2K, JPEG, WN, and BLUR. As a result, LIVE~\cite{LIVE}, CSIQ~\cite{larson:011006}, and TID2013~\cite{Ponomarenko201557} contain $634$, $600$, and $500$ test images, respectively. The Exploration database contains $4,744$ reference and $94,880$ distorted images. Although the MOS of each test image is not available in the Exploration database, innovative evaluation criteria are employed to compare BIQA measures as will be specified next.

\subsubsection{Evaluation Criteria}\label{subsubsec:ec}
We use five evaluation criteria to compare the performance of
BIQA measures. The first two are included in previous
tests carried out by the video quality experts group (VQEG)~\cite{video2000final}. Others are introduced in~\cite{ma2016waterloo} to take into account image databases without MOS. Details are given as follows.
\begin{itemize}
  \item Spearman's rank-order correlation coefficient (SRCC) is defined
    as
    \begin{equation}\label{eq:srcc}
      \mathrm{SRCC} = 1 - \frac{6\sum_i d_i^2}{N(N^2-1)}\,,
    \end{equation}
    where $N$ is the number of images in a database and $d_i$ is the difference between the $i$-th image's ranks in the MOS and model prediction.
  \item Pearson linear correlation coefficient (PLCC) is computed by
      \begin{equation}\label{eq:plcc}
          \mathrm{PLCC} = \frac{\sum_i(s_i - \bar{s})(q_i- \bar{q})}{\sqrt{\sum_i(s_i - \bar{s})^2}\sqrt{\sum_i(q_i-\bar{q})^2}}\,,
      \end{equation}
      where $s_i$ and $q_i$ stand for the MOS and model prediction of the $i$-th image, respectively.
    \item Pristine/distorted image discriminability test (D-test) considers pristine and distorted images as two distinct classes, and aims to measure how well an IQA model is able to separate the two classes. More specifically, indices of pristine and distorted images are grouped into sets $S_p$ and $S_d$, respectively. A threshold $T$ is adopted to classify images such that $S'_p = \{i|q_i>T\}$ and $S'_d = \{i|q_i\le T\}$. The average correct classification rate is defined as
        \begin{equation}\label{eq:f}
          R= \frac{1}{2}\left(\frac{|S_p\cap S'_p|}{|S_p|} + \frac{|S_d\cap S'_d|}{|S_d|}\right)\,.
        \end{equation}
        The value of $T$ should be optimized to yield the maximum correct classification rate, which results in a discriminability index
        \begin{equation}\label{eq:fopt}
          D = \max_{T} R(T)\,.
        \end{equation}
        $D$ lies in $[0,1]$ with a larger value indicating a better separability between pristine and distorted images.
    \item Listwise ranking consistency test (L-test) evaluates the robustness of IQA models when rating images with the same content and the same distortion type but different distortion levels. The  assumption is that the quality of an image degrades monotonically with the increase of the distortion level for any distortion type. Given a database with $S$ source images, $K$ distortion types and $Q$ distortion levels, the average SRCC is used to quantify the ranking consistency between distortion levels and model predictions
        \begin{equation}
         L_{s} = \frac{1}{SK}\sum_{i = 1}^{S}\sum_{j = 1}^{K}\mathrm{SRCC}({\bf l}_{ij}, {\bf q}_{ij})\,,
        \end{equation}
        where ${\bf l}_{ij}$ and ${\bf q}_{ij}$ represent the distortion levels and the corresponding distortion/quality scores given by a model to the set of images that are from the same ($i$-th) source image and have the same ($j$-th) distortion type.
    \item Pairwise preference consistency test (P-test) compares the performance of IQA models on a number of DIPs, whose generation is similar to what is described Section~\ref{subsec:DIP} but with a stricter rule~\cite{ma2016waterloo}. A good IQA model should give concordant preferences with respect to DIPs. Assuming that an image database contains $M$ DIPs and that the number of concordant pairs of an IQA model (meaning that the model predicts the correct preference) is $M_c$, the pairwise preference consistency ratio is defined as
    \begin{equation}
        P = \frac{M_c}{M}\,.
        \end{equation}
        $P$ lies in $[0,1]$ with a higher value indicating better performance. We also denote the number of incorrect preference predictions as $M_i = M- M_c $.
\end{itemize}
\begin{table}
		\centering
		\caption{Median SRCC and PLCC results across $1,000$ sessions on LIVE~\cite{LIVE}}\label{tab:1_LIVE}
	  \begin{tabular}{l|ccccc}
	      \toprule
			SRCC & JP2K & JPEG & WN & BLUR & ALL4 \\ \hline
			PSNR         & 0.908 & 0.894 & 0.984 & 0.814 & 0.883 \\
			SSIM~\cite{wang2004image} & 0.961 & 0.974 & 0.970 & 0.952 & 0.947 \\\hline
			QAC~\cite{xue2013learning}          & 0.876 & 0.951 & 0.925 & 0.911 & 0.869 \\
			NIQE~\cite{mittal2013making}         & 0.924 & 0.945 & 0.972 & {\bf 0.941} & 0.920 \\
			ILNIQE~\cite{zhang2015feature}       & 0.901 & 0.944 & {\bf 0.979} & 0.927 & 0.918 \\
			BLISS~\cite{ye2014beyond}        & 0.925 & {\bf 0.956} & 0.967 & 0.936 & 0.945 \\
			dipIQ$^*$ & {\bf 0.946}& {\bf 0.956}& {\bf 0.976} & {\bf 0.962} & {\bf 0.952}\\
            dipIQ &{\bf 0.956} & {\bf 0.969} & 0.975 & 0.940 &{\bf 0.958} \\
		\midrule
			PLCC & JP2K & JPEG & WN & BLUR & ALL4 \\\hline
			PSNR & 0.912 & 0.896 & 0.987 & 0.812 & 0.874 \\
			SSIM~\cite{wang2004image} & 0.968 & 0.980 & 0.972 & 0.951 & 0.937 \\\hline

			QAC~\cite{xue2013learning} & 0.876 & 0.960 & 0.895 & 0.912 & 0.855 \\
			NIQE~\cite{mittal2013making} & 0.932 & 0.956 & {\bf 0.979} & {\bf 0.951} & 0.912 \\
			ILNIQE~\cite{zhang2015feature} & 0.912 & 0.966 & 0.976 & 0.936 & 0.913 \\
			BLISS~\cite{ye2014beyond} & 0.933 & {\bf 0.972} & 0.978 & 0.948 & 0.945 \\
			dipIQ$^*$ &{\bf 0.958} &0.953 &0.951 &{\bf 0.950} & {\bf 0.948}\\
            dipIQ&{\bf 0.964} &{\bf 0.980} &{\bf 0.983} &0.948  &{\bf 0.957} \\
	     \bottomrule
	   \end{tabular}
	\end{table}
	
	\begin{table}[h]
		\centering
		\caption{Median SRCC and PLCC results across $1,000$ sessions on CSIQ~\cite{larson:011006}}\label{tab:1_CSIQ}
	  \begin{tabular}{l|ccccc}
	      \toprule
			SRCC & JP2K & JPEG & WN & BLUR & ALL4 \\ \hline
			PSNR & 0.941 & 0.901 & 0.943 & 0.936 & 0.928 \\
			SSIM~\cite{wang2004image} & 0.962 & 0.956 & 0.912 & 0.965 & 0.935 \\
            \hline
			QAC~\cite{xue2013learning} & 0.884 & 0.913 & 0.850 & 0.839 & 0.840 \\
			NIQE~\cite{mittal2013making} & 0.926 & 0.882 & 0.836 & 0.908 & 0.883 \\
			ILNIQE~\cite{zhang2015feature} & 0.924 & 0.905 & 0.867 & 0.867 & 0.887 \\
			BLISS~\cite{ye2014beyond} & 0.932 & {\bf 0.927} & 0.879 & 0.922 & 0.920 \\
			dipIQ$^*$ &{\bf 0.938} &0.926 & {\bf 0.887} & {\bf 0.925}& {\bf 0.924} \\
            dipIQ & {\bf 0.944} & {\bf 0.936} &{\bf 0.904}  &  {\bf 0.932}  & {\bf 0.930}\\
	\midrule
			PLCC & JP2K & JPEG & WN & BLUR & ALL4 \\ \hline
			PSNR & 0.954 & 0.908 & 0.961 & 0.937 & 0.918 \\
			SSIM~\cite{wang2004image} & 0.973 & 0.983 & 0.908 & 0.956 & 0.930 \\ \hline
			QAC~\cite{xue2013learning}& 0.898 & 0.942 & 0.865 & 0.855 & 0.847 \\
			NIQE~\cite{mittal2013making} & 0.944 & 0.946 & 0.824 & 0.935 & 0.900 \\
			ILNIQE~\cite{zhang2015feature} & 0.942 & 0.956 & 0.880 & 0.903 & 0.914 \\
			BLISS~\cite{ye2014beyond} & 0.954 & 0.970 & 0.895 & 0.947 & 0.939 \\
			dipIQ$^*$ &{\bf 0.955} &{\bf 0.971} &{\bf 0.903} &{\bf 0.951} &{\bf 0.946} \\
            dipIQ & {\bf 0.959} &{\bf 0.975}  &{\bf 0.927} &{\bf 0.958} &   {\bf 0.949}   \\
           \bottomrule
	   \end{tabular}
	\end{table}
\begin{table}
		\centering
		\caption{Median SRCC and PLCC results across $1,000$ sessions on TID2013~\cite{Ponomarenko201557}}\label{tab:1_TID013}
	  \begin{tabular}{l|ccccc}
	      \toprule
			SRCC & JP2K & JPEG & WN & BLUR & ALL4 \\ \hline
			PSNR & 0.898 & 0.929 & 0.942 & 0.965 & 0.924 \\
			SSIM~\cite{wang2004image} & 0.950 & 0.935 & 0.896 & 0.969 & 0.924 \\
            \hline
			QAC~\cite{xue2013learning} & 0.883 & 0.885 & 0.668 & 0.879 & 0.837 \\
			NIQE~\cite{mittal2013making} & 0.901 & 0.873 & 0.854 & 0.821 & 0.812 \\
			ILNIQE~\cite{zhang2015feature} & {\bf 0.912} & 0.873 & {\bf 0.890} & 0.815 & {\bf 0.881} \\
			BLISS~\cite{ye2014beyond} & 0.906 & 0.893 & 0.856 & 0.872 & 0.836 \\
			dipIQ$^*$ & 0.909 & {\bf 0.903}&0.854 &{\bf 0.884} &0.857 \\
            dipIQ &{\bf 0.926} &{\bf 0.932} &{\bf 0.905} &{\bf 0.922}  &{\bf 0.877} \\
            \midrule
			PLCC & JP2K & JPEG & WN & BLUR & ALL4 \\ \hline
			PSNR & 0.933 & 0.925 & 0.963 & 0.958 & 0.911 \\
			SSIM~\cite{wang2004image} & 0.970 & 0.968 & 0.902 & 0.958 & 0.927 \\
            \hline
			QAC~\cite{xue2013learning} & 0.892 & 0.929 & 0.719 & 0.877 & 0.829 \\
			NIQE~\cite{mittal2013making} & 0.912 & 0.928 & 0.859 & 0.848 & 0.819 \\
			ILNIQE~\cite{zhang2015feature} & 0.929 & 0.944 & {\bf 0.899} & 0.816 & 0.890 \\
			BLISS~\cite{ye2014beyond} & 0.930 & {\bf 0.963} & 0.863 & 0.872 & 0.862 \\
			dipIQ$^*$ &{\bf 0.937} &{\bf 0.963} &0.851 &{\bf 0.892} & {\bf 0.894}\\
            dipIQ &{\bf 0.948}  & {\bf 0.973} &  {\bf 0.906}  & {\bf 0.928} & {\bf 0.894} \\
		    \bottomrule
	   \end{tabular}
	\end{table}
	\begin{figure*}[t]
    \centering
    \captionsetup{justification=centering}
    
    \subfloat[]{\includegraphics[width=0.33\textwidth]{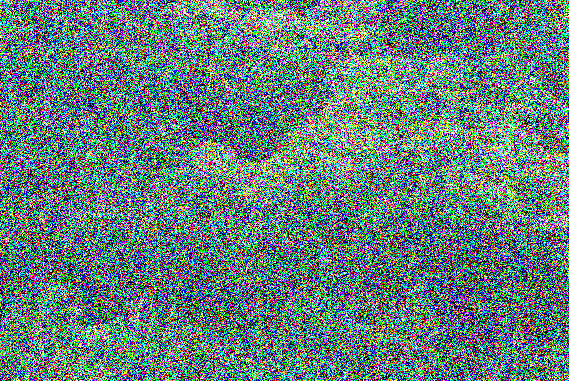}}\hskip.2em
    \subfloat[]{\includegraphics[width=0.33\textwidth]{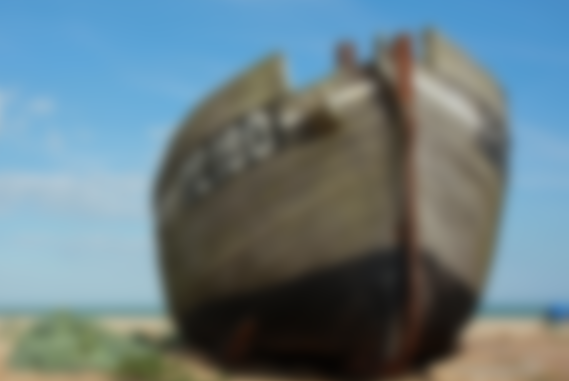}}\hskip.2em
    \subfloat[]{\includegraphics[width=0.33\textwidth]{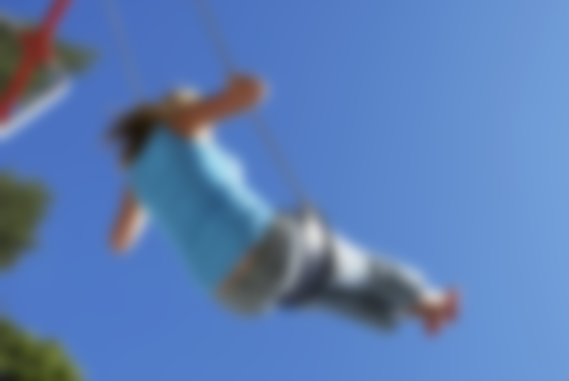}}
    \caption{The noisiness of the synthetic score~\cite{ye2014beyond}. (a) Synthetic score = 10. (b) Synthetic score = 10. (c) Synthetic score = 40.  (a)  has worse perceptual quality than (b), which in turn has approximately the same quality compared with (c). Both two cases are in disagreement with the synthetic score~\cite{ye2014beyond}. Images are selected from the training set. }\label{fig:sf}
\end{figure*}

\begin{figure*}[t!]
    \centering
    \captionsetup{justification=centering}
    
    \subfloat[]{\includegraphics[width=0.48\textwidth]{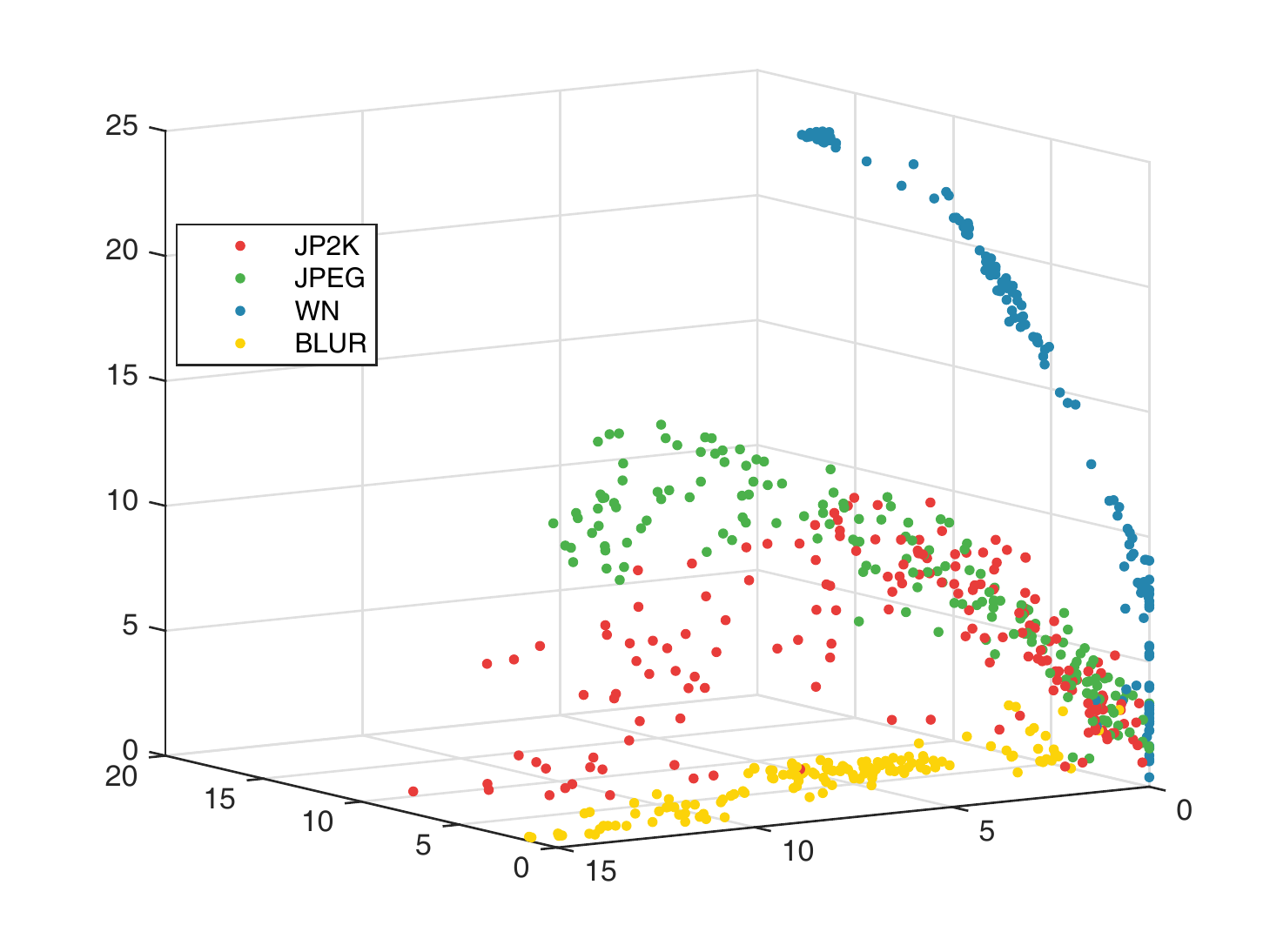}}\hskip.2em
    \subfloat[]{\includegraphics[width=0.48\textwidth]{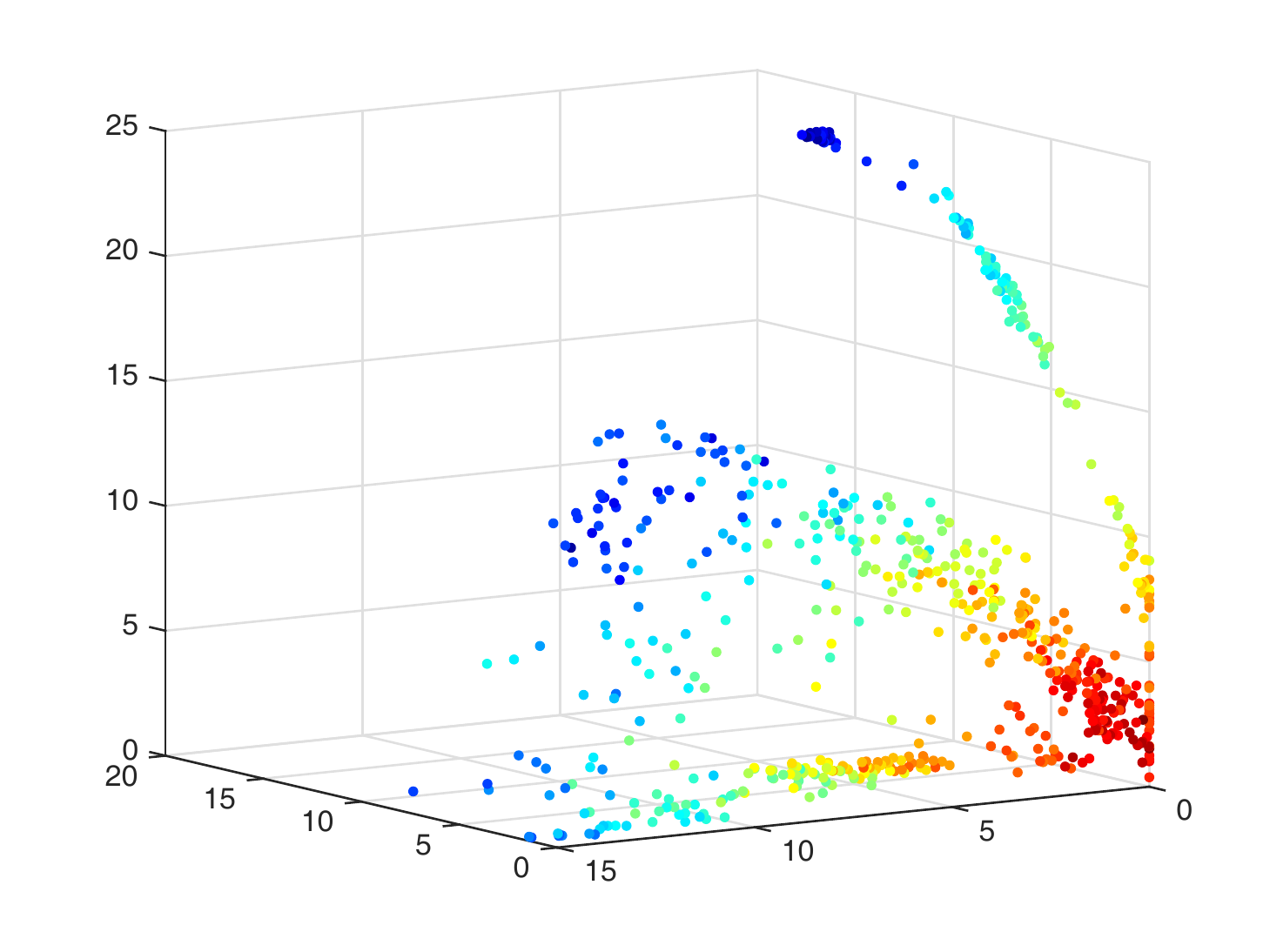}}
    \caption{Three dimensional embedding of the LIVE database~\cite{LIVE}. (a) Color encodes distortion type. (b) Color encodes quality;  the warmer, the better. The learned features from the third hidden layer of dipIQ are able to cluster images based on distortion types and align them in a perceptually meaningful way.}\label{fig:le}
\end{figure*}

SRCC and PLCC are applied to LIVE~\cite{LIVE}, CSIQ~\cite{larson:011006}, and TID2013~\cite{Ponomarenko201557}, while the D-test, L-test, and P-test are applied to the Waterloo Exploration Database. Note
that the use of PLCC requires a nonlinear function $\hat{q} = (\beta_1- \beta_2)/ (1 + \exp(-(q - \beta_3 )/|\beta_4|)) + \beta_2$ to map raw model predictions to the MOS scale. %\footnote{The logistic regression function $\hat{q} = \beta_1(\frac{1}{2} - \frac{1}{ \exp{(\beta_2 (q - \beta_3)) } } ) + \beta_4q + \beta_5$ is not used here because it cannot always preserve the monotonicity of the nonlinear mapping. Nevertheless, two nonlinear mappings give very similar median SRCC and PLCC values.}.
Following Mittal {\em et al.}~\cite{mittal2012no} and Ye {\em et al.}~\cite{ye2014beyond}, in our experiments we randomly choose $80\%$ reference images along with their corresponding distorted versions to estimate $\{\beta_i| i = 1,2,3,4\}$, and use the rest $20\%$ images for testing. This procedure is repeated $1,000$ times and the median SRCC and PLCC values are reported.
\subsection{Experimental Results}\label{subsec:er}
\subsubsection{Comparison with FR and OU-BIQA Models}
We compare dipIQ with two well-known FR-IQA models: PSNR (whose largest value is clipped at $60$ dB in order to perform a reasonable parameter estimation) and SSIM~\cite{wang2004image} (whose implementation used in the paper involves a down-sampling process~\cite{SSIMweb}) and previous OU-BIQA models, including QAC~\cite{xue2013learning}, NIQE~\cite{mittal2013making}, ILNIQE~\cite{zhang2015feature}, and BLISS~\cite{ye2014beyond}. The implementations of QAC~\cite{xue2013learning}, NIQE~\cite{mittal2013making}, and ILNIQE~\cite{zhang2015feature} are obtained from the original authors. To the best of our knowledge, the complete implementation of BLISS~\cite{ye2014beyond} is not publicly available. Therefore, to make a fair comparison we train BLISS~\cite{ye2014beyond} on the same $700$ reference images and their distorted versions, which have been used to train dipIQ. The labels are synthesized using the method in~\cite{ye2014beyond}. The training toolbox and parameter settings are inherited from  the original paper~\cite{ye2014beyond}.

		\begin{table}
		\centering
		\caption{The D-test, L-test and P-test results on the Waterloo Exploration Database~\cite{ma2016waterloo}.}\label{tab:1_Exploration}
		\begin{tabular}{l|cccc}
	      \toprule
			 & $D$ & $L_s$ & $P$ & $M_i$ \\ \hline
		PSNR &  1.0000  &  1.0000 & 0.9995  & 620,071  \\
        SSIM~\cite{wang2004image} &  1.0000  &  0.9992  &0.9991  & 1,131,457  \\\hline
        QAC~\cite{xue2013learning} &   0.9226 & 0.8699  & 0.9779  &  28,447,590 \\
        NIQE~\cite{mittal2013making}    &0.9109    &{\bf 0.9885}   & 0.9937  &  8,127,941  \\
        ILNIQE~\cite{zhang2015feature} & 0.9084   & {\bf 0.9926}  & 0.9927  &   9,435,319 \\
        BLISS~\cite{ye2014beyond} &  0.9080  &0.9801   &  0.9996 &   562,925 \\
        dipIQ$^*$& {\bf 0.9209}   & 0.9863  & {\bf 0.9996}  &  {\bf 465,069}  \\
        dipIQ & {\bf 0.9346}   & 0.9846   & {\bf 0.9999}  &{\bf 129,668}    \\		    
        \bottomrule
	   \end{tabular}
	\end{table}

	\begin{table*}
		\centering
		\caption{Statistical significance matrix based on the hypothesis testing. A symbol ``1'' means that the performance of the row algorithm is
            statistically better than that of the column algorithm, a symbol ``0'' means that the row algorithm is statistically worse, and a symbol ``-'' means
            that the row and column algorithms are statistically indistinguishable}\label{tab:staSigAna}
        \begin{tabular}{l|cccccccc}
	      \toprule
            PLCC&PSNR& SSIM &QAC  &NIQE &ILNIQE&BLISS&dipIQ$^*$&dipIQ\\\hline
        PSNR &-&0&1&0&0&0&0&0\\
        SSIM~\cite{wang2004image} &1&-&1&1&1&0&0&0\\\hline
        QAC~\cite{xue2013learning}  &0&0&-&0&0&0&0&0\\
        NIQE~\cite{mittal2013making} &1&0&1&-&-&0&0&0\\
        ILNIQE~\cite{zhang2015feature}&1&0&1&-&-&0&0&0\\
        BLISS~\cite{ye2014beyond}&1&1&1&1&1&-&0&0\\
        dipIQ$^*$&1&1&1&1&1&1&-&0\\
        dipIQ&1&1&1&1&1&1&1&-\\        
        \bottomrule
	   \end{tabular}
	\end{table*}

Tables~\ref{tab:1_LIVE},~\ref{tab:1_CSIQ}, and~\ref{tab:1_TID013} list comparison results between dipIQ and existing OU-BIQA models in terms of median SRCC and PLCC values on LIVE~\cite{LIVE}, CSIQ~\cite{larson:011006}, and TID2013~\cite{Ponomarenko201557}, respectively. Both dipIQ$^*$ and dipIQ outperform all previous OU-BIQA models on LIVE~\cite{LIVE} and CSIQ~\cite{larson:011006}, and are comparable to ILNIQE~\cite{zhang2015feature} on TID2013~\cite{Ponomarenko201557}. Although both dipIQ$^*$ and BLISS~\cite{ye2014beyond} learn a linear prediction function using CORNIA features as inputs~\cite{ye2012unsupervised}, we observe consistent performance gains of dipIQ$^*$ across all three databases over BLISS~\cite{ye2014beyond}. This may be because dipIQ$^*$ learns from more reliable data (DIPs) with uncertainty weighting, whereas the training labels (synthetic scores) for BLISS are noisier, as exemplified in Fig.~\ref{fig:sf}. It is not hard to observe that Fig.~\ref{fig:sf}(a) has clearly worse perceptual quality than Fig.~\ref{fig:sf}(b), which in turn has approximately the same quality compared with Fig.~\ref{fig:sf}(c). Both two cases are in disagreement with the synthetic score~\cite{ye2014beyond}.

To ascertain that the improvement of dipIQ is statistically significant, we carry out a two sample T-test (with a $95\%$ confidence) between PLCC values obtained by different models on LIVE~\cite{LIVE}. After comparing every possible pairs of OU-BIQA models, the results are summarized in Table~\ref{tab:staSigAna}, where a symbol ``1'' means the row model performs significantly better than the column model, a symbol ``0'' means the opposite, and a symbol ``-'' indicates that the row and column models are statistically indistinguishable. It can be observed that dipIQ is statistically better than dipIQ$^*$, which is better than all previous OU-BIQA models.

Table~\ref{tab:1_Exploration} shows the results on the Waterloo Exploration Database. dipIQ$^*$ and dipIQ outperform all previous OU-BIQA models in the D-test and P-test, and are competitive in the L-test, whose performance is slightly inferior to NIQE~\cite{mittal2013making} and ILNIQE~\cite{zhang2015feature}. By learning from examples with a variety of image content, dipIQ is able to crush the number of incorrect preference predictions in the P-test down to around $130,000$ out of more than $1$ billion candidate DIPs.

In order to gain intuitions on why the generalizability of dipIQ is excellent even without MOS for training, we visualize the three-dimensional embedding of the LIVE database~\cite{LIVE} in Fig~\ref{fig:le}, using the learned three-dimensional features from the third hidden layer of dipIQ. We can see that the learned representation is able to cluster test images according to the distortion type, and meanwhile align them with respect to their perceptual quality in a meaningful way, where high quality images are clamped together regardless of image content.

\begin{table}[t]
		\centering
		\caption{Median SRCC and PLCC results across $1,000$ sessions, training on LIVE~\cite{LIVE} and testing on CSIQ~\cite{larson:011006}. The superscripts $B$ and $D$ indicate that the input features of dipIQ are from BRISQUE~\cite{mittal2012no} and DIIVINE~\cite{moorthy2011blind}, respectively}\label{tab:2_CSIQ}
		\begin{tabular}{l|ccccc}
	      \toprule
			SRCC & JP2K & JPEG & WN & BLUR & ALL4 \\ \hline
			BRISQUE~\cite{mittal2012no} & 0.894 & 0.916 & {\bf 0.934} & 0.915 & 0.909 \\
			dipIQ$^{B}$& {\bf 0.938}&   {\bf0.938}&  {\bf0.934}& {\bf 0.943}  &{\bf0.926}\\ 

			DIIVINE~\cite{moorthy2011blind} & 0.844 & 0.819 & 0.881 & 0.884 & 0.835 \\
			dipIQ$^{D}$ &  {\bf 0.930}    &{\bf 0.939}    &{\bf 0.904}    &{\bf 0.920}& {\bf 0.912}\\ 
			CORNIA~\cite{ye2012unsupervised} & 0.916 & 0.919 & 0.787 & 0.928 & 0.915 \\
dipIQ & {\bf 0.944} & {\bf 0.936} &{\bf 0.904}  &  {\bf 0.932}  & {\bf 0.930}\\
	\midrule
			PLCC & JP2K & JPEG & WN & BLUR & ALL4 \\ \hline
			BRISQUE~\cite{mittal2012no} & 0.937 & 0.960 & {\bf 0.947} & 0.936 & 0.937 \\
			dipIQ$^{B}$ &{\bf 0.956} &{\bf0.974} & 0.945 &{\bf 0.959} & {\bf 0.943}\\ 

			DIIVINE~\cite{moorthy2011blind} & 0.898 & 0.818 & 0.903 & 0.909 & 0.855 \\
			dipIQ$^{D}$ & {\bf 0.949 }  & {\bf 0.973}  &  {\bf 0.924} &   {\bf 0.944} &  {\bf 0.942}\\ 
			CORNIA~\cite{ye2012unsupervised} & 0.947 & 0.960 & 0.777 & 0.953 & 0.934 \\
            dipIQ & {\bf 0.959} &{\bf 0.975}  &{\bf 0.927} &{\bf 0.958} &   {\bf 0.949}   \\
		\bottomrule
		\end{tabular}
	\end{table}
	
	\begin{table}[t]
		\centering
		\caption{Median SRCC and PLCC results across $1,000$ sessions, training on LIVE~\cite{LIVE} and testing on TID2013~\cite{Ponomarenko201557}}\label{tab:2_TID2013}
		\begin{tabular}{l|ccccc}
	      \toprule
			SRCC & JP2K & JPEG & WN & BLUR & ALL4 \\ \hline
			BRISQUE~\cite{mittal2012no} & 0.906 & 0.894 & 0.889 & 0.886 & {\bf0.883} \\
			dipIQ$^{B}$ &{\bf 0.927 }&  {\bf 0.921}&  {\bf 0.921} & {\bf 0.917} & {\bf 0.883}\\ 

			DIIVINE~\cite{moorthy2011blind} & 0.857 & 0.680 & 0.879 & 0.859 & 0.795 \\

			dipIQ$^{D}$ & {\bf 0.912} &{\bf 0.889} &{\bf 0.887 }  & {\bf 0.905}  &{\bf 0.872}\\ 
			CORNIA~\cite{ye2012unsupervised} & 0.907 & 0.912 & 0.798 &{\bf  0.934}&{\bf 0.893} \\
            dipIQ &{\bf 0.926} &{\bf 0.932} &{\bf 0.905} &0.922 & 0.877 \\
            \midrule
			PLCC & JP2K & JPEG & WN & BLUR & ALL4 \\ \hline
			BRISQUE~\cite{mittal2012no} & 0.919 & 0.950 & 0.886 & 0.884 & {\bf 0.901} \\
			dipIQ$^{B}$ &{\bf 0.942}  & {\bf0.957}&  {\bf0.923}&  {\bf 0.906}& 0.883\\ 

			DIIVINE~\cite{moorthy2011blind} & 0.901 & 0.696 & {\bf 0.882} & 0.860 & 0.794 \\
            dipIQ$^{D}$ &{\bf 0.945}    &{\bf 0.947}&    0.881  &  {\bf 0.896} &  {\bf  0.892}\\ 
			CORNIA~\cite{ye2012unsupervised} & 0.923 & 0.960 & 0.778 & {\bf 0.934} & {\bf 0.904} \\
            dipIQ &{\bf 0.948}  & {\bf 0.973} &  {\bf 0.906}  &  0.928 & 0.894 \\	
            	\bottomrule
		\end{tabular}
	\end{table}

\begin{table}[t]
		\centering
		\caption{The D-test, L-test and P-test results on the Exploration database~\cite{ma2016waterloo}, training on LIVE~\cite{LIVE}}\label{tab:2_Exploration}
		\begin{tabular}{l|cccc}
	      \toprule
			 & $D$ & $L_s$ & $P$ & $M_i$ \\ \hline
		BRISQUE~\cite{mittal2012no} &  0.9204 &  {\bf 0.9772} & 0.9930  & 9,004,685  \\
        dipIQ$^B$ & {\bf 0.9265}  & 0.9753  & {\bf 0.9996}& {\bf 503,911} \\
        DIIVINE~\cite{moorthy2011blind} &   0.8538 & 0.8908  & 0.9540 &  59,053,011 \\
        dipIQ$^D$ & {\bf 0.9191} & {\bf 0.9588} & {\bf 0.9983} & {\bf 2,124,199}\\
        CORNIA~\cite{ye2012unsupervised} & 0.9290  & 0.9764 & 0.9947  &   6,808,400 \\
        dipIQ & {\bf 0.9346}   & {\bf 0.9846}   & {\bf 0.9999}  &{\bf 129,668}    \\            	
        \bottomrule
		\end{tabular}
	\end{table}

\begin{figure}[]
    \centering
    \captionsetup{justification=centering}
    \subfloat[]{\includegraphics[width=0.24\textwidth]{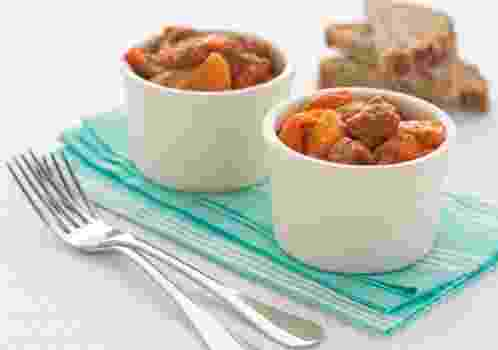}}\hskip.2em
    \subfloat[]{\includegraphics[width=0.24\textwidth]{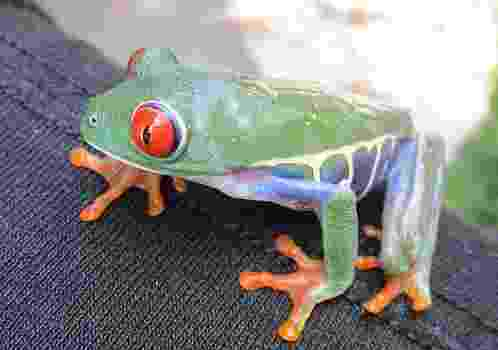}}\hskip.2em
    \vspace{2pt}
    \subfloat[]{\includegraphics[width=0.24\textwidth]{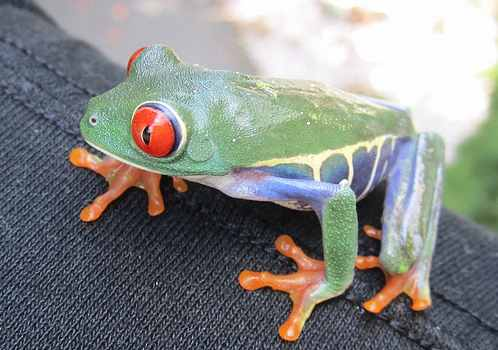}}\hskip.2em
    \subfloat[]{\includegraphics[width=0.24\textwidth]{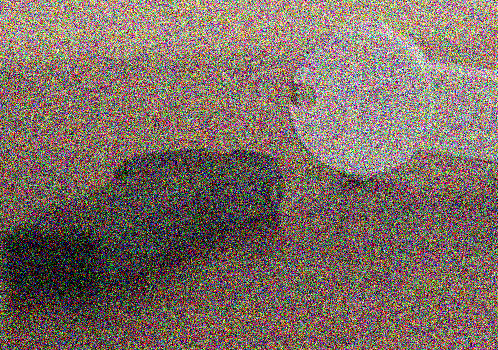}}
    \caption{gMAD competition between dipIQ$^B$  and BRISQUE~\cite{mittal2012no}. (a) best BRISQUE for fixed dipIQ$^B$. (b) worst BRISQUE for fixed dipIQ$^B$. (c) best dipIQ$^B$ for fixed BRISQUE. (d) worst dipIQ$^B$ for fixed BRISQUE.}\label{fig:gmad}
\end{figure}

\subsubsection{Comparison with OA-BIQA Models}
In the second set of experiments, we train dipIQ using different feature representations as inputs and compare with OA-BIQA models using the same representations and MOS for training. BRISQUE~\cite{mittal2012no} and DIIVINE~\cite{moorthy2011blind} are selected as representative features extracted from the spatial and wavelet domain, respectively. We also compare dipIQ with CORNIA~\cite{ye2012unsupervised}, whose features are adopted as the default input to dipIQ. We re-train BRISQUE~\cite{mittal2012no}, DIIVINE~\cite{moorthy2011blind}, and CORNIA~\cite{ye2012unsupervised} on the LIVE database, whose learning tools and  parameter settings follow their respective papers. We adjust the dimension of the input layer of dipIQ to accommodate features of different dimensions and train them on the $700$ reference images and their distorted versions, as described in~\ref{subsec:id}. All models are tested on CSIQ~\cite{larson:011006}, TID2013~\cite{Ponomarenko201557} and the Exportation database~\cite{ma2016waterloo}. From Tables~\ref{tab:2_CSIQ},~\ref{tab:2_TID2013}, and~\ref{tab:2_Exploration}, we observe that dipIQ consistently performs better than the corresponding OA-BIQA model on CSIQ~\cite{larson:011006} and the Exploration database, and is comparable on TID2013~\cite{Ponomarenko201557}. The reason we do not obtain noticeable  performance gains on TID2013~\cite{Ponomarenko201557} may be that TID2013~\cite{Ponomarenko201557} has $18$ references images originated from
LIVE~\cite{LIVE}, based on which the OA-BIQA models have been trained. This creates dependencies between training and testing sets. We may also draw conclusions about the effectiveness of the feature representations based on their performance under the same pairwise L2R framework: generally speaking, CORNIA~\cite{ye2012unsupervised} features $>$ BRISQUE~\cite{mittal2012no} features $>$ DIIVINE~\cite{moorthy2011blind} features.

We further compare dipIQ$^B$ and BRISQUE~\cite{mittal2012no} using the gMAD competition methodology on the Waterloo Exploration Database. Specifically, we first find a pair of images that have the maximum and minimum dipIQ$^B$ values from a subset of images in the Exploration database, where BRISQUE~\cite{mittal2012no} rates them to have the same quality. We then repeat this procedure, but with the roles of  dipIQ$^B$  and BRISQUE~\cite{mittal2012no} exchanged. The two image pairs are shown in Fig.~\ref{fig:gmad}, from which we conclude that images in the first row exhibits approximately the same
perceptual quality (in agreement with dipIQ$^B$) and those
in the second row has drastically different perceptual quality
(in disagreement with BRISQUE~\cite{mittal2012no}). This verifies that the robustness of  dipIQ$^B$ is significantly improved over BRISQUE~\cite{mittal2012no} using the same feature representations and MOS for training. Similar gMAD competition results are obtained across all quality levels, and for dipIQ$^D$ versus DIIVINE~\cite{moorthy2011blind} and dipIQ versus CORNIA~\cite{ye2012unsupervised}.

In summary, the proposed pairwise L2R approach is proved to learn OU-BIQA models with improved generalizability and robustness compared with OA-BIQA models using the same feature representations and MOS for training.

\section{Listwise L2R Approach for OU-BIQA}\label{sec:ListNet}
In this section, we extend the proposed pairwise L2R approach for OU-BIQA to a listwise L2R one. Specifically, we first construct three-element DILs by concatenating DIPs. For example, given two DIPs $\langle i,j\rangle$ and $\langle j,k\rangle $ with the same level of uncertainty, we create a list $\langle i,j,k\rangle$ with the ground truth label $\bar{P}_{ijk} = 1$,  indicating that the quality of the $i$-th image is better than the $j$-th image, whose quality is better than the $k$-th image. The uncertainty level is transferred as well. We then employ ListNet~\cite{cao2007learning}, a listwise L2R extension of RankNet~\cite{Burges05learningto} to learn OU-BIQA models. The major differences between ListNet and RankNet are twofold. First, ListNet can have multiple streams with the same weights to accommodate a list of inputs, where each stream is implemented by a classical neural network architecture similar to RankNet, as shown in Fig.~\ref{fig:ranknet}. In this paper, we instantiate a three-stream ListNet to fit three-element DILs. Second, the loss function of ListNet is defined using the concept of permutation probability. More specifically, we define a permutation $\pi = \langle \pi(1),\pi(2),\ldots,\pi(n)\rangle $ on a list of $n$ instances as a bijection from $\{1,2,..,n\}$ to itself, where $\pi(j)$ denotes the instance at position $j$ in the permutation. The set of all possible permutations of $n$ instances is termed as $\Pi$. We define the probability of permutation $\pi$ given the list
of predicted scores $\{f({\bf x}_i)\}$ as
\begin{equation}\label{eq:pp}
P_{\pi}(f) = \prod_{j = 1}^{n}\frac{\exp\left(f({\bf x}_{\pi(j)})\right)}{\sum\limits_{k = j}\limits^n\exp\left(f({\bf x}_{\pi(k)})\right)}\,,
\end{equation}
which satisfies $P_{\pi}(f) >0$ and $\sum_{\pi \in \Pi}P_{\pi}(f) = 1$ as proved in~\cite{cao2007learning}. The loss function can then be defined as the cross entropy function between the ground truth and permutation probabilities
\begin{equation}\label{eq:lce}
L(f;\{{\bf x}_i\},\{\bar{P}_{\pi}\}) = -\sum_{\pi \in \Pi}\bar{P}_{\pi}\log(P_{\pi})\,.
\end{equation}
When $n = 2$, the loss function of ListNet~\cite{cao2007learning} in Eq.~(\ref{eq:lce}) becomes equivalent to that of RankNet~\cite{Burges05learningto} in Eq.~(\ref{eq:ce}). In the case of three-element DILs, we have $\bar{P}_{\pi} = 1$, if ${\pi = \langle i,j,k\rangle}$ and $\bar{P}_{\pi} = 0$ otherwise. Therefore, the loss function in Eq.~(\ref{eq:lce}) can be simplified as
\begin{equation}\label{eq:lces}
\begin{split}
& L(f;{\bf x}_i,{\bf x}_j,{\bf x}_k,\bar{P}_{ ijk }) = -f({\bf x}_i) - f({\bf x}_j)\\
         &+ \log\left(\sum_{l\in\{i,j,k\}}\exp\left(f({\bf x}_l)\right)\right)+\log\left(\sum_{l\in\{j,k\}}\exp\left(f({\bf x}_l)\right)\right)\,,
\end{split}
\end{equation}
base on which we define the batch-level loss as
\begin{equation}\label{eq:minilosslist}
L_b(f) = \sum_{\langle i, j, k\rangle \in \mathcal{B}}(1 - U_{ ijk})L(f;{\bf x}_i,{\bf x}_j, {\bf x}_k, \bar{P}_{ijk})\,,
\end{equation}
where $U_{ ijk}$ is the uncertainty level of the list, transferred from the corresponding DIPs.  The gradient of Eq.~(\ref{eq:minilosslist}) w.r.t. the parameters $\bf w$ can be easily derived. Note that ListNet~\cite{cao2007learning} does not add new parameters.

We generate $50$ million DILs from the available DIPs as the training data for ListNet~\cite{cao2007learning}. The training procedure is exactly the same as training RankNet~\cite{Burges05learningto}. The training stops
when the entire set of image lists have been swept once. The weights that achieve the lowest validation set loss are used for testing.

We list the comparison results between dilIQ trained by ListNet~\cite{cao2007learning} and the baseline dipIQ on LIVE~\cite{LIVE}, CSIQ~\cite{larson:011006}, TID2013~\cite{Ponomarenko201557}, and the Exploration database in Tables~\ref{tab:3_LIVE},~\ref{tab:3_CSIQ},~\ref{tab:3_TID2013}, and~\ref{tab:3_Exploration}, respectively. Remarkable performance improvements have been achieved on CSIQ and TID2013. This may be because the ranking position information is made explicit to the learning process. dilIQ is comparable to dipIQ on LIVE and the Exploration database.


\begin{table}[t]
		\centering
		\caption{Median SRCC and PLCC results across $1,000$ sessions on LIVE~\cite{LIVE}, using ListNet~\cite{cao2007learning} for training}\label{tab:3_LIVE}
		\begin{tabular}{l|ccccc}
	      \toprule
			 SRCC& JP2K & JPEG & WN &BLUR & ALL \\ \hline
            dipIQ &0.956 & 0.969 & 0.975 & 0.940 &0.958 \\
        %dipIQ^I$ & 0.954   & 0.966 &   {\bf 0.976} &   {\bf 0.950}   & {\bf 0.961}\\
        dilIQ &  0.956  &  0.966 &   {\bf 0.976}  &  {\bf 0.953}  &  0.958 \\
        \midrule
			 PLCC& JP2K & JPEG & WN &BLUR & ALL \\ \hline
            dipIQ&0.964 &0.980 &0.983 &0.948  &0.957 \\
       % dipIQ$^I$ &  0.962   & 0.975   & 0.977  &  {\bf 0.949}  &  0.957\\
        dilIQ &  0.964  &  0.978   & {\bf 0.985}  &  {\bf 0.956} &   0.954 \\        
        \bottomrule
		\end{tabular}
	\end{table}

\begin{table}[t]
		\centering
		\caption{Median SRCC and PLCC results across $1,000$ sessions on CSIQ~\cite{larson:011006}, using ListNet~\cite{cao2007learning} for training}\label{tab:3_CSIQ}
		\begin{tabular}{l|ccccc}
	      \toprule
			 SRCC& JP2K & JPEG & WN &BLUR & ALL \\ \hline
            dipIQ & 0.944 &  0.936 &0.904  &   0.932  &  0.930\\
       % dipIQ$^I$ & 0.940  &  {\bf 0.945} &   0.903 &   {\bf 0.934} &   {\bf 0.934} \\
        dilIQ & 0.930 &   0.925   & 0.893   & {\bf 0.939} &   {\bf 0.936} \\
        \midrule
			 PLCC& JP2K & JPEG & WN &BLUR & ALL \\ \hline
            dipIQ & 0.959 &0.975 &0.927 &0.958&   0.949\\
        %dipIQ$^I$ &0.956 &   {\bf 0.977} &   0.916 &   0.955 &   {\bf 0.951} \\
        dilIQ &  0.954 &   0.968   & 0.920  &  {\bf 0.960}  &  {\bf 0.954}\\
                \bottomrule
		\end{tabular}
	\end{table}

\begin{table}[t]
		\centering
		\caption{Median SRCC and PLCC results across $1,000$ sessions on TID2013~\cite{Ponomarenko201557}, using ListNet~\cite{cao2007learning} for training}\label{tab:3_TID2013}
		\begin{tabular}{l|ccccc}
	      \toprule
			 SRCC& JP2K & JPEG & WN &BLUR & ALL \\ \hline
            dipIQ &0.926 &0.932&0.905 & 0.922  &0.877 \\
        %dipIQ$^I$ & 0.922  &  0.929 &   0.897 &   0.922  &  {\bf 0.883}  \\
        dilIQ &  0.918   & 0.849  &  0.905  &  {\bf 0.925}  &  {\bf 0.891}\\
        \midrule
			 PLCC& JP2K & JPEG & WN &BLUR & ALL \\ \hline
        dipIQ & 0.948  &  0.973 &   0.906  &  0.928 &  0.894 \\
        %dipIQ$^I$ &  {\bf 0.951}   & 0.971&    0.895 &   0.922 &   {\bf 0.907} \\
        dilIQ &  0.948&    0.923   & 0.903   & {\bf 0.929 } &  {\bf 0.915} \\
                \bottomrule
		\end{tabular}
	\end{table}

\begin{table}[t]
		\centering
		\caption{The D-test, L-test and P-test results on the Exploration database~\cite{ma2016waterloo}, using ListNet~\cite{cao2007learning} for training}\label{tab:3_Exploration}
		\begin{tabular}{l|cccc}
	      \toprule
			 & $D$ & $L_s$ & $P$ & $M_i$ \\ \hline
        dipIQ &  0.9346   & 0.9846   &  0.9999  & 129,668    \\
        %dipIQ$^I$ &{\bf 0.9398}  & {\bf 0.9847}& {\bf 0.9999}&{\bf 118,136}  \\
        dilIQ & 0.9346  & {\bf 0.9893} & 0.9998& 198,650\\
         \bottomrule
		\end{tabular}
	\end{table}

\section{Conclusion and Future Work}\label{sec:con}
In this paper, we have proposed an OU-BIQA model, namely dipIQ, using RankNet~\cite{Burges05learningto}.  The input to the dipIQ training model are an enormous number of DIPs, not obtained by expensive subjective testing but automatically generated with the help of most trusted FR-IQA models at low cost. Extensive experimental results demonstrate the effectiveness of the proposed dipIQ indices with higher accuracy and improved robustness in content variations. We also learn an OU-BIQA model, namely dilIQ, using a listwise L2R approach, which achieves an additional performance gain.

The current work opens the door to a new class of OU-BIQA models and can be extended in many ways. First, novel image pair and list generation engines may be developed to account for situations that reference images are not available (or do not ever exist). Second, advanced L2R algorithms are worth exploring to improve the quality prediction performance. Third, in practice, a pair of images may be regarded as having indiscriminable quality. Such knowledge could be obtained either from subjective testing (\textit{e.g.}, paired comparison between images) or from the image source (\textit{e.g.}, two pristine images acquired from the same source), and is informative in constraining the behavior of an objective quality model. The current learning framework needs to be improved in order to learn from such quality-indiscriminable image pairs. Fourth, given the powerful DIP generation engine developed in the current work and the remarkable success of recent deep convolutional neural networks, it may become feasible to develop end-to-end BIQA models that bypass the feature extraction process and achieve even stronger robustness and generalizability.

\section*{Acknowledgment}
The authors would like to thank Zhengfang Duanmu for suggestions on the efficient implementation of RankNet, and the anonymous reviewers for constructive comments. This work was supported in part by the Natural Sciences and Engineering Research Council of Canada, and the Australian Research Council Projects FT-130101457, DP-140102164, and LP-150100671. K. Ma was partially supported by the CSC.
\ifCLASSOPTIONcaptionsoff
  \newpage
\fi



%\bibliographystyle{IEEEtran}

\bibliographystyle{IEEEtran}
\bibliography{Kede,egbibshort}

\begin{thebibliography}{10}
\providecommand{\url}[1]{#1}
\csname url@samestyle\endcsname
\providecommand{\newblock}{\relax}
\providecommand{\bibinfo}[2]{#2}
\providecommand{\BIBentrySTDinterwordspacing}{\spaceskip=0pt\relax}
\providecommand{\BIBentryALTinterwordstretchfactor}{4}
\providecommand{\BIBentryALTinterwordspacing}{\spaceskip=\fontdimen2\font plus
\BIBentryALTinterwordstretchfactor\fontdimen3\font minus
  \fontdimen4\font\relax}
\providecommand{\BIBforeignlanguage}[2]{{%
\expandafter\ifx\csname l@#1\endcsname\relax
\typeout{** WARNING: IEEEtran.bst: No hyphenation pattern has been}%
\typeout{** loaded for the language `#1'. Using the pattern for}%
\typeout{** the default language instead.}%
\else
\language=\csname l@#1\endcsname
\fi
#2}}
\providecommand{\BIBdecl}{\relax}
\BIBdecl

\bibitem{wu2005digital}
H.~R. Wu and K.~R. Rao, \emph{Digital Video Image Quality and Perceptual
  Coding}.\hskip 1em plus 0.5em minus 0.4em\relax CRC press, 2005.

\bibitem{wang2006modern}
Z.~Wang and A.~C. Bovik, \emph{Modern Image Quality Assessment}.\hskip 1em plus
  0.5em minus 0.4em\relax Morgan \& Claypool Publishers, 2006.

\bibitem{daly1992visible}
S.~J. Daly, ``Visible differences predictor: An algorithm for the assessment of
  image fidelity,'' in \emph{SPIE/IS\&T Symposium on Electronic Imaging:
  Science and Technology}, 1992, pp. 2--15.

\bibitem{wang2006quality}
Z.~Wang, G.~Wu, H.~R. Sheikh, E.~P. Simoncelli, E.-H. Yang, and A.~C. Bovik,
  ``Quality-aware images,'' \emph{IEEE Transactions on Image Processing},
  vol.~15, no.~6, pp. 1680--1689, Jun. 2006.

\bibitem{wang2011reduced}
Z.~Wang and A.~C. Bovik, ``Reduced- and no-reference image quality assessment:
  The natural scene statistic model approach,'' \emph{IEEE Signal Processing
  Magazine}, vol.~28, no.~6, pp. 29--40, Nov. 2011.

\bibitem{moorthy2010two}
A.~K. Moorthy and A.~C. Bovik, ``A two-step framework for constructing blind
  image quality indices,'' \emph{IEEE Signal Processing Letters}, vol.~17,
  no.~5, pp. 513--516, May 2010.

\bibitem{saad2012blind}
M.~A. Saad, A.~C. Bovik, and C.~Charrier, ``Blind image quality assessment: A
  natural scene statistics approach in the {DCT} domain,'' \emph{IEEE
  Transactions on Image Processing}, vol.~21, no.~8, pp. 3339--3352, Aug. 2012.

\bibitem{mittal2012no}
A.~Mittal, A.~K. Moorthy, and A.~C. Bovik, ``No-reference image quality
  assessment in the spatial domain,'' \emph{IEEE Transactions on Image
  Processing}, vol.~21, no.~12, pp. 4695--4708, Dec. 2012.

\bibitem{ye2012unsupervised}
P.~Ye, J.~Kumar, L.~Kang, and D.~Doermann, ``Unsupervised feature learning
  framework for no-reference image quality assessment,'' in \emph{IEEE
  Conference on Computer Vision and Pattern Recognition}, 2012, pp. 1098--1105.

\bibitem{moorthy2011blind}
A.~K. Moorthy and A.~C. Bovik, ``Blind image quality assessment: From natural
  scene statistics to perceptual quality,'' \emph{IEEE Transactions on Image
  Processing}, vol.~20, no.~12, pp. 3350--3364, Dec. 2011.

\bibitem{wu2015a}
Q.~Wu, Z.~Wang, and H.~Li, ``A highly efficient method for blind image quality
  assessment,'' in \emph{IEEE International Conference on Image Processing},
  2015, pp. 339--343.

\bibitem{xue2014blind}
W.~Xue, X.~Mou, L.~Zhang, A.~C. Bovik, and X.~Feng, ``Blind image quality
  assessment using joint statistics of gradient magnitude and {L}aplacian
  features,'' \emph{IEEE Transactions on Image Processing}, vol.~23, no.~11,
  pp. 4850--4862, Nov. 2014.

\bibitem{gu2015using}
K.~Gu, G.~Zhai, X.~Yang, and W.~Zhang, ``Using free energy principle for blind
  image quality assessment,'' \emph{IEEE Transactions on Multimedia}, vol.~17,
  no.~1, pp. 50--63, Jan. 2015.

\bibitem{wu2014blind}
Q.~Wu, H.~Li, F.~Meng, K.~N. Ngan, B.~Luo, C.~Huang, and B.~Zeng, ``Blind image
  quality assessment based on multi-channel features fusion and label
  transfer,'' \emph{IEEE Transactions on Circuits and Systems for Video
  Technology}, vol.~26, no.~3, pp. 425--440, Mar. 2016.

\bibitem{Ponomarenko201557}
N.~Ponomarenko, L.~Jin, O.~Ieremeiev, V.~Lukin, K.~Egiazarian, J.~Astola,
  B.~Vozel, K.~Chehdi, M.~Carli, F.~Battisti, and C.-C.~J. Kuo, ``Image
  database {TID2013}: Peculiarities, results and perspectives,'' \emph{Signal
  Processing: Image Communication}, vol.~30, pp. 57--77, Jan. 2015.

\bibitem{Burges05learningto}
C.~Burges, T.~Shaked, E.~Renshaw, A.~Lazier, M.~Deeds, N.~Hamilton, and
  G.~Hullender, ``Learning to rank using gradient descent,'' in
  \emph{International Conference on Machine Learning}, 2005, pp. 89--96.

\bibitem{liu2009learning}
T.-Y. Liu, ``Learning to rank for information retrieval,'' \emph{Foundations
  and Trends in Information Retrieval}, vol.~3, no.~3, pp. 225--331, 2009.

\bibitem{hang2011short}
L.~Hang, ``A short introduction to learning to rank,'' \emph{IEICE Transactions
  on Information and Systems}, vol.~94, no.~10, pp. 1854--1862, Oct. 2011.

\bibitem{ma2016group}
K.~Ma, Q.~Wu, Z.~Wang, Z.~Duanmu, H.~Yong, H.~Li, and L.~Zhang, ``Group {MAD}
  competition $-$ a new methodology to compare objective image quality
  models,'' in \emph{IEEE Conference on Computer Vsion and Pattern
  Recognition}, 2016, pp. 1664--1673.

\bibitem{ma2016waterloo}
K.~Ma, Z.~Duanmu, Q.~Wu, Z.~Wang, H.~Yong, H.~Li, and L.~Zhang, ``{Waterloo
  Exploration Database}: New challenges for image quality assessment models,''
  \emph{IEEE Transactions on Image Processing}, vol.~26, no.~2, pp. 1004--1016,
  Feb. 2017.

\bibitem{cao2007learning}
Z.~Cao, T.~Qin, T.-Y. Liu, M.-F. Tsai, and H.~Li, ``Learning to rank: From
  pairwise approach to listwise approach,'' in \emph{International Conference
  on Machine Learning}, 2007, pp. 129--136.

\bibitem{wu1997generalized}
H.~R. Wu and M.~Yuen, ``A generalized block-edge impairment metric for video
  coding,'' \emph{IEEE Signal Processing Letters}, vol.~4, no.~11, pp.
  317--320, Nov. 1997.

\bibitem{Wang2000Blind}
Z.~Wang, A.~C. Bovik, and B.~L. Evan, ``Blind measurement of blocking artifacts
  in images,'' in \emph{IEEE International Conference on Image Processing},
  2000, pp. 981--984.

\bibitem{liu2002efficient}
S.~Liu and A.~C. Bovik, ``Efficient {DCT}-domain blind measurement and
  reduction of blocking artifacts,'' \emph{IEEE Transactions on Circuits and
  Systems for Video Technology}, vol.~12, no.~12, pp. 1139--1149, Dec. 2002.

\bibitem{tong2004blur}
H.~Tong, M.~Li, H.~Zhang, and C.~Zhang, ``Blur detection for digital images
  using wavelet transform,'' in \emph{IEEE International Conference on
  Multimedia and Expo}, 2004, pp. 17--20.

\bibitem{wang2003local}
Z.~Wang and E.~P. Simoncelli, ``Local phase coherence and the perception of
  blur,'' in \emph{Advances in Neural Information Processing Systems}, 2003.

\bibitem{zhu2009no}
X.~Zhu and P.~Milanfar, ``A no-reference sharpness metric sensitive to blur and
  noise,'' in \emph{International Workshop on Quality of Multimedia
  Experience}, 2009, pp. 64--69.

\bibitem{ouguz1998image}
S.~O{\u{g}}uz, Y.~Hu, and T.~Q. Nguyen, ``Image coding ringing artifact
  reduction using morphological post-filtering,'' in \emph{IEEE Workshop on
  Multimedia Signal Processing}, 1998, pp. 628--633.

\bibitem{sheikh2005no}
H.~R. Sheikh, A.~C. Bovik, and L.~Cormack, ``No-reference quality assessment
  using natural scene statistics: {JPEG2000},'' \emph{IEEE Transactions on
  Image Processing}, vol.~14, no.~11, pp. 1918--1927, Nov. 2005.

\bibitem{liu2010no}
H.~Liu, N.~Klomp, and I.~Heynderickx, ``A no-reference metric for perceived
  ringing artifacts in images,'' \emph{IEEE Transactions on Circuits and
  Systems for Video Technology}, vol.~20, no.~4, pp. 529--539, Apr. 2010.

\bibitem{wandell1995foundations}
B.~A. Wandell, \emph{Foundations of Vision.}\hskip 1em plus 0.5em minus
  0.4em\relax Sinauer Associates, 1995.

\bibitem{hubel1962receptive}
D.~H. Hubel and T.~N. Wiesel, ``Receptive fields, binocular interaction and
  functional architecture in the cat's visual cortex,'' \emph{The Journal of
  physiology}, vol. 160, no.~1, pp. 106--154, Jan. 1962.

\bibitem{heeger1992normalization}
D.~J. Heeger, ``Normalization of cell responses in cat striate cortex,''
  \emph{Visual Neuroscience}, vol.~9, no.~02, pp. 181--197, Aug. 1992.

\bibitem{field1994goal}
D.~J. Field, ``What is the goal of sensory coding?'' \emph{Neural Computation},
  vol.~6, no.~4, pp. 559--601, Jul. 1994.

\bibitem{geisler2002bayesian}
W.~S. Geisler and R.~L. Diehl, ``Bayesian natural selection and the evolution
  of perceptual systems,'' \emph{Philosophical Transactions of the Royal
  Society of London B: Biological Sciences}, vol. 357, no. 1420, pp. 419--448,
  Apr. 2002.

\bibitem{simoncelli1992shiftable}
E.~P. Simoncelli, W.~T. Freeman, E.~H. Adelson, and D.~J. Heeger, ``Shiftable
  multiscale transforms,'' \emph{IEEE Transactions on Information Theory},
  vol.~38, no.~2, pp. 587--607, Mar. 1992.

\bibitem{mallat1989theory}
S.~G. Mallat, ``A theory for multiresolution signal decomposition: The wavelet
  representation,'' \emph{IEEE Transactions on Pattern Analysis and Machine
  Intelligence}, vol.~11, no.~7, pp. 674--693, Jul. 1989.

\bibitem{li2002blind}
X.~Li, ``Blind image quality assessment,'' in \emph{IEEE International
  Conference on Image Processing}, 2002, pp. 449--452.

\bibitem{marziliano2004perceptual}
P.~Marziliano, F.~Dufaux, S.~Winkler, and T.~Ebrahimi, ``Perceptual blur and
  ringing metrics: Application to {JPEG2000},'' \emph{Signal Processing: Image
  Communication}, vol.~19, no.~2, pp. 163--172, Feb. 2004.

\bibitem{li2011blind}
C.~Li, A.~C. Bovik, and X.~Wu, ``Blind image quality assessment using a general
  regression neural network,'' \emph{IEEE Transactions on Neural Networks},
  vol.~22, no.~5, pp. 793--799, May 2011.

\bibitem{fang2015no}
Y.~Fang, K.~Ma, Z.~Wang, W.~Lin, and G.~Zhai, ``No-reference quality assessment
  of contrast-distorted images based on natural scene statistics,'' \emph{IEEE
  Signal Processing Letters}, vol.~22, no.~7, pp. 838--842, Jul. 2015.

\bibitem{zhu2010automatic}
X.~Zhu and P.~Milanfar, ``Automatic parameter selection for denoising
  algorithms using a no-reference measure of image content,'' \emph{IEEE
  Transactions on Image Processing}, vol.~19, no.~12, pp. 3116--3132, Dec.
  2010.

\bibitem{mittal2013making}
A.~Mittal, R.~Soundararajan, and A.~C. Bovik, ``Making a ``completely blind''
  image quality analyzer,'' \emph{IEEE Signal Processing Letters}, vol.~20,
  no.~3, pp. 209--212, Mar. 2013.

\bibitem{mittal2012blind}
A.~Mittal, G.~S. Muralidhar, J.~Ghosh, and A.~C. Bovik, ``Blind image quality
  assessment without human training using latent quality factors,'' \emph{IEEE
  Signal Processing Letters}, vol.~19, no.~2, pp. 75--78, Feb. 2012.

\bibitem{ye2013real}
P.~Ye, J.~Kumar, L.~Kang, and D.~Doermann, ``Real-time no-reference image
  quality assessment based on filter learning,'' in \emph{IEEE Conference on
  Computer Vision and Pattern Recognition}, 2013, pp. 987--994.

\bibitem{wang2005reduced}
Z.~Wang and E.~P. Simoncelli, ``Reduced-reference image quality assessment
  using a wavelet-domain natural image statistic model,'' in \emph{Human Vision
  and Electronic Imaging}, 2005, pp. 149--159.

\bibitem{hou2015blind}
W.~Hou, X.~Gao, D.~Tao, and X.~Li, ``Blind image quality assessment via deep
  learning,'' \emph{IEEE Transactions on Neural Networks and Learning Systems},
  vol.~26, no.~6, pp. 1275--1286, Jun. 2015.

\bibitem{li2009reduced}
Q.~Li and Z.~Wang, ``Reduced-reference image quality assessment using divisive
  normalization-based image representation,'' \emph{IEEE Journal of Selected
  Topics in Signal Processing}, vol.~3, no.~2, pp. 202--211, Apr. 2009.

\bibitem{rehman2012reduced}
A.~Rehman and Z.~Wang, ``Reduced-reference image quality assessment by
  structural similarity estimation,'' \emph{IEEE Transactions on Image
  Processing}, vol.~21, no.~8, pp. 3378--3389, Aug. 2012.

\bibitem{tang2011learning}
H.~Tang, N.~Joshi, and A.~Kapoor, ``Learning a blind measure of perceptual
  image quality,'' in \emph{IEEE Conference on Computer Vision and Pattern
  Recognition}, 2011, pp. 305--312.

\bibitem{tang2014blind}
------, ``Blind image quality assessment using semi-supervised rectifier
  networks,'' in \emph{IEEE Conference on Computer Vision and Pattern
  Recognition}, 2014, pp. 2877--2884.

\bibitem{ye2012no}
P.~Ye and D.~Doermann, ``No-reference image quality assessment using visual
  codebooks,'' \emph{IEEE Transactions on Image Processing}, vol.~21, no.~7,
  pp. 3129--3138, Jul. 2012.

\bibitem{hassen2013image}
R.~Hassen, Z.~Wang, and M.~M. Salama, ``Image sharpness assessment based on
  local phase coherence,'' \emph{IEEE Transactions on Image Processing},
  vol.~22, no.~7, pp. 2798--2810, Jul. 2013.

\bibitem{xu2010two}
L.~Xu and J.~Jia, ``Two-phase kernel estimation for robust motion deblurring,''
  in \emph{European Conference on Computer Vision}, 2010, pp. 157--170.

\bibitem{wang2002no}
Z.~Wang, H.~R. Sheikh, and A.~C. Bovik, ``No-reference perceptual quality
  assessment of {JPEG} compressed images,'' in \emph{IEEE International
  Conference on Image Processing}, vol.~1, 2002, pp. 477--480.

\bibitem{huang1975importance}
T.~S. Huang, J.~W. Burnett, and A.~G. Deczky, ``The importance of phase in
  image processing filters,'' \emph{IEEE Transactions on Acoustics, Speech and
  Signal Processing}, vol.~23, no.~6, pp. 529--542, Dec. 1975.

\bibitem{oppenheim1981importance}
A.~V. Oppenheim and J.~S. Lim, ``The importance of phase in signals,''
  \emph{Proceedings of the IEEE}, vol.~69, no.~5, pp. 529--541, May 1981.

\bibitem{kovesi1999image}
P.~Kovesi, ``Image features from phase congruency,'' \emph{Journal of Computer
  Vision Research}, vol.~1, no.~3, pp. 1--26, Jun. 1999.

\bibitem{saad2010dct}
M.~A. Saad, A.~C. Bovik, and C.~Charrier, ``A {DCT} statistics-based blind
  image quality index,'' \emph{IEEE Signal Processing Letters}, vol.~17, no.~6,
  pp. 583--586, Jun. 2010.

\bibitem{xue2013learning}
W.~Xue, L.~Zhang, and X.~Mou, ``Learning without human scores for blind image
  quality assessment,'' in \emph{IEEE Conference on Computer Vision and Pattern
  Recognition}, 2013, pp. 995--1002.

\bibitem{zhang2011fsim}
L.~Zhang, L.~Zhang, X.~Mou, and D.~Zhang, ``{FSIM}: A feature similarity index
  for image quality assessment,'' \emph{IEEE Transactions on Image Processing},
  vol.~20, no.~8, pp. 2378--2386, Aug. 2011.

\bibitem{kang2014convolutional}
L.~Kang, P.~Ye, Y.~Li, and D.~Doermann, ``Convolutional neural networks for
  no-reference image quality assessment,'' in \emph{IEEE Conference on Computer
  Vision and Pattern Recognition}, 2014, pp. 1733--1740.

\bibitem{cortes1995support}
C.~Cortes and V.~Vapnik, ``Support-vector networks,'' \emph{Machine Learning},
  vol.~20, no.~3, pp. 273--297, Sep. 1995.

\bibitem{scholkopf2000new}
B.~Sch{\"o}lkopf, A.~J. Smola, R.~C. Williamson, and P.~L. Bartlett, ``New
  support vector algorithms,'' \emph{Neural Computation}, vol.~12, no.~5, pp.
  1207--1245, May 2000.

\bibitem{zhang2015feature}
L.~Zhang, L.~Zhang, and A.~Bovik, ``A feature-enriched completely blind image
  quality evaluator,'' \emph{IEEE Transactions on Image Processing}, vol.~24,
  no.~8, pp. 2579--2591, Aug. 2015.

\bibitem{xu2014rank}
L.~Xu, W.~Lin, J.~Li, X.~Wang, Y.~Yan, and Y.~Fang, ``Rank learning on training
  set selection and image quality assessment,'' in \emph{IEEE International
  Conference on Multimedia and Expo}, 2014, pp. 1--6.

\bibitem{gao2015learning}
F.~Gao, D.~Tao, X.~Gao, and X.~Li, ``Learning to rank for blind image quality
  assessment,'' \emph{IEEE Transactions on Neural Networks and Learning
  Systems}, vol.~26, no.~10, pp. 2275--2290, Oct. 2015.

\bibitem{hofmann2001unsupervised}
T.~Hofmann, ``Unsupervised learning by probabilistic latent semantic
  analysis,'' \emph{Machine Learning}, vol.~42, no.~1, pp. 177--196, Jan. 2001.

\bibitem{gao2013universal}
X.~Gao, F.~Gao, D.~Tao, and X.~Li, ``Universal blind image quality assessment
  metrics via natural scene statistics and multiple kernel learning,''
  \emph{IEEE Transactions on Neural Networks and Learning Systems}, vol.~24,
  no.~12, pp. 2013--2026, Dec. 2013.

\bibitem{Fuhr1989Optimum}
N.~Fuhr, ``Optimum polynomial retrieval functions based on the probability
  ranking principle,'' \emph{ACM Transactions on Information Systems}, vol.~7,
  no.~3, pp. 183--204, Jul. 1989.

\bibitem{Cossock2006Subset}
D.~Cossock and T.~Zhang, ``Subset ranking using regression,'' in
  \emph{Conference on Learning Theory}, 2006, pp. 605--619.

\bibitem{Nallapati2004Discriminative}
R.~Nallapati, ``Discriminative models for information retrieval,'' in
  \emph{International ACM SIGIR Conference on Research and Development in
  Information Retrieval}, 2004, pp. 64--71.

\bibitem{Crammer2002Pranking}
K.~Crammer and Y.~Singer, ``Pranking with ranking,'' in \emph{Advances in
  Neural Information Processing Systems}, 2002, pp. 641--647.

\bibitem{shashua2002ranking}
A.~Shashua and A.~Levin, ``Ranking with large margin principle: Two
  approaches,'' in \emph{Advances in Neural Information Processing Systems},
  2002, pp. 937--944.

\bibitem{Joachims2002Optimizing}
T.~Joachims, ``Optimizing search engines using clickthrough data,'' in
  \emph{Eighth ACM SIGKDD International Conference on Knowledge Discovery and
  Data Mining}, 2002, pp. 133--142.

\bibitem{Tsai2007FRank}
M.~F. Tsai, T.~Y. Liu, T.~Qin, H.~H. Chen, and W.~Y. Ma, ``{FRank}: A ranking
  method with fidelity loss,'' in \emph{International ACM SIGIR Conference on
  Research and Development in Information Retrieval}, 2007, pp. 383--390.

\bibitem{Freund2003An}
Y.~Freund, R.~Iyer, R.~E. Schapire, and Y.~Singer, ``An efficient boosting
  algorithm for combining preferences,'' \emph{Journal of Machine Learning
  Research}, vol.~4, no.~6, pp. 170--178, Nov. 2003.

\bibitem{Freund1995A}
Y.~Freund and R.~E. Schapire, ``A decision-theoretic generalization of online
  learning and an application to boosting,'' in \emph{European Conference on
  Computational Learning Theory}, 1995, pp. 23--37.

\bibitem{taylor2008softrank}
M.~Taylor, J.~Guiver, S.~Robertson, and T.~Minka, ``{SoftRank}: optimizing
  non-smooth rank metrics,'' in \emph{ACM International Conference on Web
  Search and Data Mining}, 2008, pp. 77--86.

\bibitem{yue2007support}
Y.~Yue, T.~Finley, F.~Radlinski, and T.~Joachims, ``A support vector method for
  optimizing average precision,'' in \emph{International ACM SIGIR Conference
  on Research and Development in Information Retrieval}, 2007, pp. 271--278.

\bibitem{yeh2007learning}
J.-Y. Yeh, J.-Y. Lin, H.-R. Ke, and W.-P. Yang, ``Learning to rank for
  information retrieval using genetic programming,'' in \emph{SIGIR Workshop on
  Learning to Rank for Information Retrieval}, 2007, pp. 1--8.

\bibitem{wang2003multiscale}
Z.~Wang, E.~P. Simoncelli, and A.~C. Bovik, ``Multiscale structural similarity
  for image quality assessment,'' in \emph{IEEE Asilomar Conference on Signals,
  Systems and Computers}, 2003, pp. 1398--1402.

\bibitem{sheikh2006image}
H.~R. Sheikh and A.~C. Bovik, ``Image information and visual quality,''
  \emph{IEEE Transactions on Image Processing}, vol.~15, no.~2, pp. 430--444,
  Feb. 2006.

\bibitem{xue2014gradient}
W.~Xue, L.~Zhang, X.~Mou, and A.~C. Bovik, ``Gradient magnitude similarity
  deviation: A highly efficient perceptual image quality index,'' \emph{IEEE
  Transactions on Image Processing}, vol.~23, no.~2, pp. 684--695, Feb. 2014.

\bibitem{sheikh2006statistical}
H.~R. Sheikh, M.~F. Sabir, and A.~C. Bovik, ``A statistical evaluation of
  recent full reference image quality assessment algorithms,'' \emph{IEEE
  Transactions on Image Processing}, vol.~15, no.~11, pp. 3440--3451, Nov.
  2006.

\bibitem{LIVE}
H.~R. Sheikh, Z.~Wang, A.~C. Bovik, and L.~K. Cormack, {Image} and Video
  Quality Assessment Research at \text{LIVE} [Online]. Available:
  http://live.ece.utexas.edu/research/quality/.

\bibitem{hinton2006fast}
G.~E. Hinton, S.~Osindero, and Y.-W. Teh, ``A fast learning algorithm for deep
  belief nets,'' \emph{Neural Computation}, vol.~18, no.~7, pp. 1527--1554,
  Jul. 2006.

\bibitem{krizhevsky2012imagenet}
A.~Krizhevsky, I.~Sutskever, and G.~E. Hinton, ``{ImageNet} classification with
  deep convolutional neural networks,'' in \emph{Advances in Neural Information
  Processing Systems}, 2012, pp. 1097--1105.

\bibitem{ye2014beyond}
P.~Ye, J.~Kumar, and D.~Doermann, ``Beyond human opinion scores: blind image
  quality assessment based on synthetic scores,'' in \emph{IEEE Conference on
  Computer Vision and Pattern Recognition}, 2014, pp. 4241--4248.

\bibitem{nair2010rectified}
V.~Nair and G.~E. Hinton, ``Rectified linear units improve restricted boltzmann
  machines,'' in \emph{IEEE International Conference on Machine Learning},
  2010, pp. 807--814.

\bibitem{Simonyan2015Very}
K.~Simonyan and A.~Zisserman, ``Very deep convolutional networks for
  large-scale image recognition,'' in \emph{International Conference on
  Learning Representation}, 2015.

\bibitem{larson:011006}
E.~C. Larson and D.~M. Chandler, ``Most apparent distortion: Full-reference
  image quality assessment and the role of strategy,'' \emph{SPIE Journal of
  Electronic Imaging}, vol.~19, no.~1, pp. 1--21, Jan. 2010.

\bibitem{video2000final}
VQEG, {Final} Report from the Video Quality Experts Group on the Validation of
  Objective Models of Video Quality Assessment 2000 [Online]. Available:
  http://www.vqeg.org.

\bibitem{wang2004image}
Z.~Wang, A.~C. Bovik, H.~R. Sheikh, and E.~P. Simoncelli, ``Image quality
  assessment: From error visibility to structural similarity,'' \emph{IEEE
  Transactions on Image Processing}, vol.~13, no.~4, pp. 600--612, Apr. 2004.

\bibitem{SSIMweb}
------, {The} SSIM Index for Image Quality Assessment [Online]. Available:
  \url{https://ece.uwaterloo.ca/~z70wang/research/ssim/}.

\end{thebibliography}

\begin{IEEEbiography}[{\includegraphics[width=1in,height=1.25in,clip,keepaspectratio]{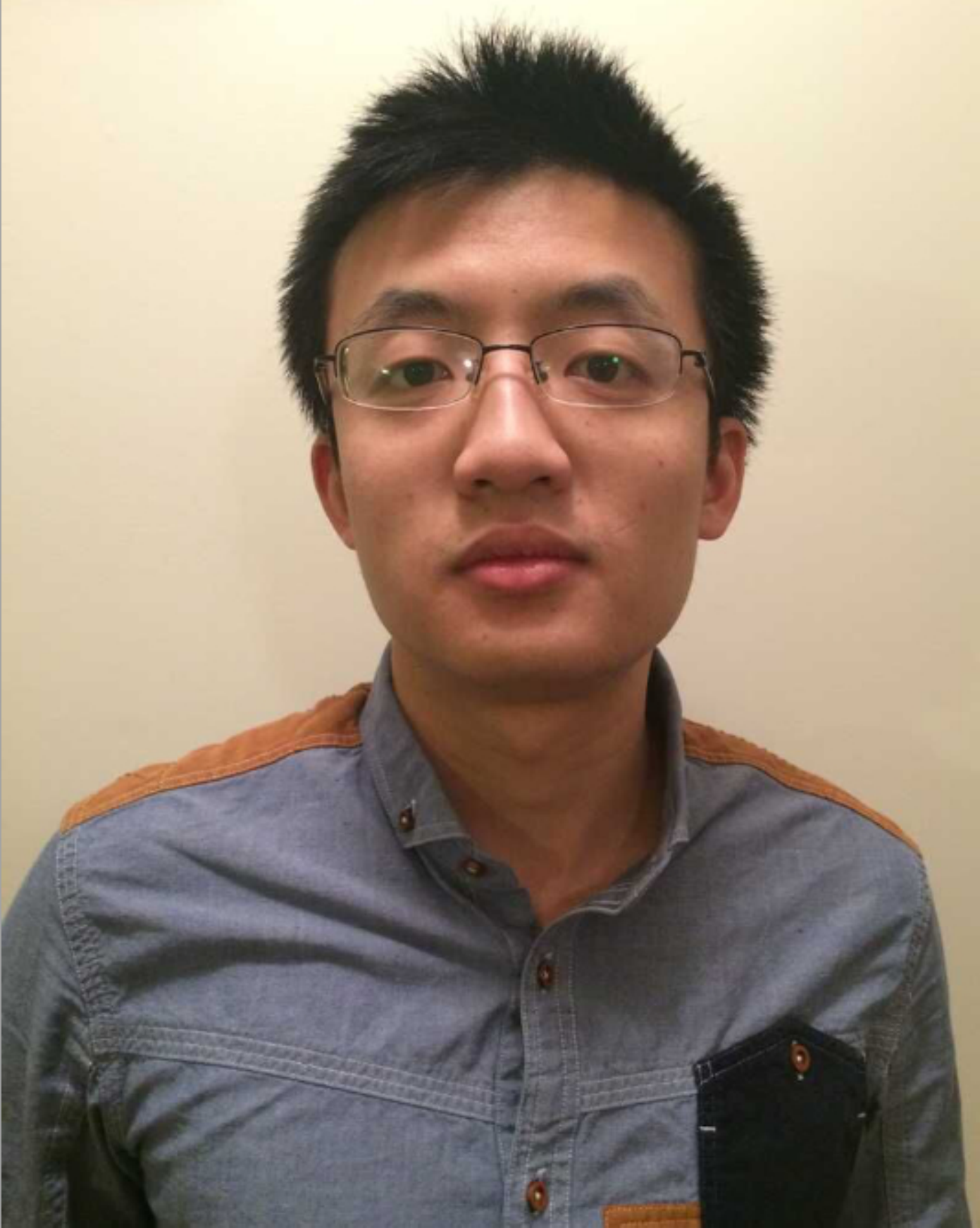}}]{Kede Ma}
(S'13) received the B.E. degree from University of Science and Technology of China, Hefei, China, in 2012 and M.A.Sc. degree from University of Waterloo, ON, Canada, where he is currently working toward the Ph.D. degree in electrical and computer engineering. His research interests lie in perceptual image processing and computational photography.
\end{IEEEbiography}

\begin{IEEEbiography}[{\includegraphics[width=1in,height=1.25in,clip,keepaspectratio]{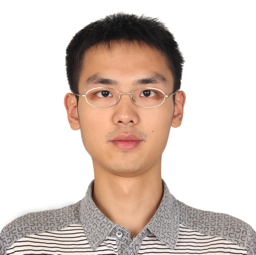}}]{Wentao Liu}
(S'15) received the B.E. and the M.E. degrees from Tsinghua University, Beijing, China in 2011 and 2014, respectively. He is currently working toward the Ph.D. degree in the Electrical \& Computer Engineering Department, University of Waterloo, ON, Canada. His research interests include perceptual quality assessment of images and videos.
\end{IEEEbiography}

\begin{IEEEbiography}[{\includegraphics[width=1in,height=1.25in,clip,keepaspectratio]{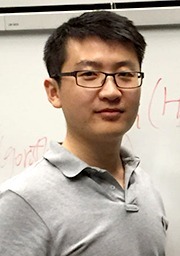}}]{Tongliang Liu}
is currently a Lecturer with the School of Information Technologies and the Faculty of Engineering and Information Technologies, and a core member in the UBTech Sydney AI Institute, at The University of Sydney. He received the BEng degree in electronic engineering and information science from the University of Science and Technology of China, and the PhD degree from the University of Technology Sydney. His research interests include statistical learning theory, computer vision, and optimization. He has authored and co-authored 20+ research papers including IEEE T-PAMI, T-NNLS, T-IP, ICML, and KDD.
\end{IEEEbiography}

\begin{IEEEbiography}[{\includegraphics[width=1in,height=1.25in,clip,keepaspectratio]{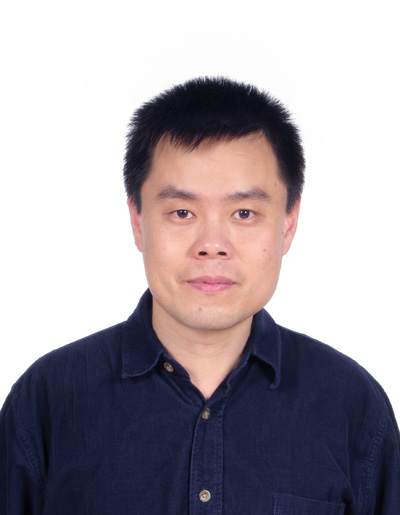}}]{Zhou Wang}
(S'99-M'02-SM'12-F'14) received the Ph.D. degree from The University of Texas at Austin in 2001. He is currently a Professor in the Department of Electrical and Computer Engineering, University of Waterloo, Canada. His research interests include image processing, coding, and quality assessment; computational vision and pattern analysis; multimedia communications; and biomedical signal processing. He has more than 100 publications in these fields with over 30,000 citations (Google Scholar).

Dr. Wang serves as a Senior Area Editor of IEEE Transactions on Image Processing (2015-present), and an Associate Editor of IEEE Transactions on Circuits and Systems for Video Technology (2016-present). Previously, he served as a member of IEEE Multimedia Signal Processing Technical Committee (2013-2015), an Associate Editor of IEEE Transactions on Image Processing (2009-2014), Pattern Recognition (2006-present) and IEEE Signal Processing Letters (2006-2010), and a Guest Editor of IEEE Journal of Selected Topics in Signal Processing (2013-2014 and 2007-2009). He is a Fellow of Canadian Academy of Engineering, and a recipient of 2016 IEEE Signal Processing Society Sustained Impact Paper Award, 2015 Primetime Engineering Emmy Award, 2014 NSERC E.W.R. Steacie Memorial Fellowship Award, 2013 IEEE Signal Processing Magazine Best Paper Award, 2009 IEEE Signal Processing Society Best Paper Award, and 2009 Ontario Early Researcher Award.
\end{IEEEbiography}

\begin{IEEEbiography}[{\includegraphics[width=1in,height=1.25in,clip,keepaspectratio]{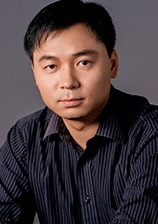}}]{Dacheng Tao}
(F'15) is Professor of Computer Science and ARC Future Fellow in the School of Information Technologies and the Faculty of Engineering and Information Technologies, and the Inaugural Director of the UBTech Sydney Artificial Intelligence Institute, at The University of Sydney. He mainly applies statistics and mathematics to Artificial Intelligence and Data Science. His research interests spread across computer vision, data science, image processing, machine learning, and video surveillance. His research results have expounded in one monograph and 500+ publications at prestigious journals and prominent conferences, such as IEEE T-PAMI, T-NNLS, T-IP, JMLR, IJCV, NIPS, CIKM, ICML, CVPR, ICCV, ECCV, AISTATS, ICDM; and ACM SIGKDD, with several best paper awards, such as the best theory/algorithm paper runner up award in IEEE ICDM'07, the best student paper award in IEEE ICDM'13, the 2014 ICDM 10-year highest-impact paper award, and the 2017 IEEE Signal Processing Society Best Paper Award. He received the 2015 Australian Scopus-Eureka Prize, the 2015 ACS Gold Disruptor Award and the 2015 UTS Vice-Chancellor's Medal for Exceptional Research. He is a Fellow of the IEEE, OSA, IAPR and SPIE.
\end{IEEEbiography}


%\end{thebibliography}


%\end{IEEEbiography}


%\newpage



%\vfill


\end{document}